\pdfoutput=1
\documentclass[10pt,twocolumn,letterpaper]{article}

\usepackage{cvpr}              

\usepackage[table]{xcolor}
\usepackage{graphicx}
\usepackage{amsmath}
\usepackage{amssymb}
\usepackage{booktabs}

\usepackage{hyperref}

\usepackage[capitalize]{cleveref}

\crefname{section}{Sec.}{Secs.}
\Crefname{section}{Section}{Sections}
\Crefname{table}{Table}{Tables}
\crefname{table}{Tab.}{Tabs.}

\usepackage{comment}

\usepackage{float}
\usepackage{array}
\usepackage{multirow}
\usepackage{footnote}
\usepackage[accsupp]{axessibility}

\makeatletter
\@namedef{ver@everyshi.sty}{}
\makeatother
\usepackage{tikz}
\usepackage{pgfplots}
\pgfplotsset{compat=1.15}
\usepgfplotslibrary{colorbrewer}

\def\etal{\emph{et al.}}
\def\ie{\emph{i.e.}}
\def\eg{\emph{e.g.}}

\def\R{\mathbb{R}}
\def\xm{q_{\max}}
\def\sdsdata{SDS-ST1k}
\newcommand{\floor}[1]{\left\lfloor{#1}\right\rfloor}
\newcolumntype{S}{>{\centering\arraybackslash}m{0.9cm}}
\newcolumntype{M}{>{\centering\arraybackslash}m{1.1cm}}
\newcolumntype{L}{>{\centering\arraybackslash}m{1.8cm}}

\newcommand{\trule}{\specialrule{\arrayrulewidth}{5mm}{0mm}}
\newcommand{\mruleb}{\specialrule{\arrayrulewidth}{0mm}{0mm}}
\newcommand{\mruleu}{\specialrule{\arrayrulewidth}{0mm}{0mm}}
\newcommand{\brule}{\specialrule{\arrayrulewidth}{0mm}{5mm}}

\def\cvprPaperID{42}
\def\confName{CVPRW}
\def\confYear{2022}

\begin{document}
\title{A Low Memory Footprint Quantized Neural Network for \\ Depth Completion of Very Sparse Time-of-Flight Depth Maps}
\author{
    Xiaowen Jiang\textsuperscript{1}
    \quad
    Valerio Cambareri\textsuperscript{2}\thanks{Corresponding author:~\url{valerio.cambareri@sony.com}. XJ, CU, and AS thank Sony Europe B.V. for supporting their internships.}
    \quad
    Gianluca Agresti\textsuperscript{3}
    \\
    Cynthia Ifeyinwa Ugwu\textsuperscript{4}
    \quad
    Adriano Simonetto\textsuperscript{4}
    \quad
    Fabien Cardinaux\textsuperscript{3}
    \quad
    Pietro Zanuttigh\textsuperscript{4}\vspace{0.5ex} \\ 
	\textsuperscript{1} EPFL, Switzerland
    \quad \textsuperscript{2} Sony Depthsensing Solutions NV, Belgium\\
    \quad \textsuperscript{3} Sony Europe B.V., R\&D Center, Stuttgart Laboratory 1, Germany
    \quad \textsuperscript{4} University of Padova, Italy
}
\maketitle
\vspace{-10mm}
\begin{abstract}
\vspace{-2mm}
Sparse active illumination enables precise time-of-flight depth sensing as it maximizes signal-to-noise ratio for low power budgets. However, depth completion is required to produce dense depth maps for 3D perception. 
We address this task with realistic illumination and sensor resolution constraints by simulating ToF datasets for indoor 3D perception with challenging sparsity levels. 
We propose a quantized convolutional encoder-decoder network for this task. Our model achieves optimal depth map quality by means of input pre-processing and carefully tuned training with a geometry-preserving loss function. 
We also achieve low memory footprint for weights and activations by means of mixed precision quantization-at-training techniques. 
The resulting quantized models are comparable to the state of the art in terms of quality, but they require very low GPU times and achieve up to $14$-fold memory size reduction for the weights w.r.t. their floating point counterpart with minimal impact on quality metrics.
\end{abstract}
\setlength{\belowcaptionskip}{1mm}
\setlength{\abovecaptionskip}{1mm}
\setlength{\abovedisplayskip}{4pt}
\setlength{\belowdisplayskip}{4pt}
\renewcommand{\arraystretch}{1.25}
\vspace{-4mm}
\section{Introduction}
\label{sec:intro}
Time-of-flight (ToF) sensors are active depth sensing devices~\cite{zanuttigh_time--flight_2016} with the potential to provide more reliable scene understanding by true 3D perception. Due to their low power consumption and accuracy at real-time frame rates, ToF sensors were recently integrated in mobile consumer devices~\cite{luetzenburg_evaluation_2021}. 
However, ToF relies on active illumination~\cite{zanuttigh_time--flight_2016} which accounts for a relevant part of its power consumption. To use the limited power budget of a mobile device more efficiently, the scene can be illuminated with a dot pattern light source so that its radiant intensity concentrates onto a small number of regions (\emph{dots}). 
A low-power ToF sensor for indoors 3D perception typically captures $500 \sim 1500$ dots per frame. Because of this sparsity level, sensor fusion techniques are necessary to obtain dense depth maps. Hereafter, we tackle this \emph{depth completion} task by using
a single RGB image frame together with a sparse depth map warped to the RGB camera view (at small baseline of a few $mm$ to minimize occlusion issues w.r.t. the ToF sensor). We dub this the \emph{sparse ToF} setting. 

To obtain dense depth from a sparse depth map, deep learning-based approaches have gained significant attention, with several contributions showing competitive performances especially on autonomous driving datasets \cite{uhrig_sparsity_2017, sun_scalability_2020, guizilini_3d_2020}. We instead focus on efficient indoor 3D perception with ToF. In order to attain overall low power budget for both ToF sensing and depth completion, neural networks must be designed and trained accordingly for efficient inference. In this paper we introduce a carefully designed encoder-decoder CNN for depth completion. We achieve a low memory footprint for weights and activations by means of mixed-precision quantization-aware training and show that the resulting quantized models achieve conspicuous memory size savings for weights and activations, while remaining competitive w.r.t. quality metrics. Our contributions are as follows:
{
    \begin{itemize}
        \setlength{\itemsep}{1pt}
        \setlength{\parskip}{0pt}
        \setlength{\parsep}{0pt}
        \item We propose an efficient depth completion model suitable for low-power sparse ToF sensing. We evaluate our model and other state-of-the-art models on two datasets: the public-domain dataset NYU-Depth v2 \cite{silberman_indoor_2012}, and \sdsdata~\cite{anon_sds-st1k_2021}, a contributed dataset providing accurate sparse ToF and RGB camera simulations.
        \item We analyze the impact of input pre-processing providing fast initialization; a normals estimation block enabling supervision with ground truth normals; different loss functions and network configurations. Each of these elements contributes to achieving optimal depth map quality for the resulting model.
        \item We evaluate quantization-aware training strategies to quantize weights and activations of the previous model. To the best of our knowledge, mixed-precision quantization~\cite{uhlich_mixed_2020} has been used successfully to classification tasks. In this paper we present the first application of mixed precision quantization to the depth completion task. This technique allows us to minimize the weights and activations memory requirements with very limited quality degradation.
    \end{itemize}
}

\section{Related Works}
\label{sec:prior}
\subsection{Depth Completion}
The main task we explore in this paper is popular in computer vision since the introduction of depth sensors~\cite{diebel_application_2005, barron_2016} and has been addressed by neural networks with convolutional encoder-decoder architectures~\cite{ma_sparse--dense_2018, jaritz_sparse_2018, ma_self-supervised_2019, qiu_deeplidar_2019, cheng_learning_2019, yang_dense_2019, park_non-local_2020}. Many other works define the state of the art in depth completion~\cite{imran_depth_2019, eldesokey_confidence_2020, eldesokey_uncertainty-aware_2020,  li_multi-scale_2020, lopez-rodriguez_project_2020, qu_depth_2020, teixeira_aerial_2020, hu_penet_2021}. We will compare our results against CNN solutions~\cite{chen_estimating_2018, ma_self-supervised_2019, cheng_learning_2019, park_non-local_2020} selected on the basis of their similar complexity (single encoder-decoder). These will be retrained on our sparse ToF datasets. There exist also higher-complexity networks leveraging, \eg, multiple decoder branches~\cite{xu_depth_2019, qiu_deeplidar_2019} or separate encoder branches~\cite{yang_dense_2019, van_gansbeke_sparse_2019, shivakumar_dfusenet_2019, guizilini_sparse_2021} but in this work we aimed for a more compact and lightweight model. 
We will only focus on \emph{single-frame} depth completion
. Many contributions extend depth completion by either self-supervision or by providing as input multiple RGB-D frames with known or inferred poses~\cite{ma_self-supervised_2019, wong_unsupervised_2020, patil_dont_2020, guizilini_sparse_2021, wimbauer_monorec_2021}. By tackling only the single-frame case, we grant temporally-independent behavior for both dynamic and static cameras and remove additional computational effort required by pose estimation via neural networks (\eg, \cite{kendall_posenet_2015}) or traditional odometry. We now discuss the main works we compare against.
\vspace{-4mm}
\paragraph{Sparse-to-Dense (S2D~\cite{ma_self-supervised_2019})}
is a popular approach using a single encoder-decoder network. Its core contribution is self-supervision by pose estimation (a feature we do not leverage) as well as its long-standing role in the KITTI leaderboard compared to its low complexity. 
The network architecture is similar to ours
. As shown in~\cite[Sec.~6.4]{ma_self-supervised_2019} and confirmed in our experiments, its robustness to sparsity is limited. We reproduced and retrained their model.
\vspace{-4mm}
\paragraph{$\boldsymbol{\rm D}^{\boldsymbol 3}$~\cite{chen_estimating_2018}}
is developed using DenseNet~\cite{Huang_2017_CVPR} layers as building blocks. Its input pre-processing is an efficient way to initialize the depth map and attain high quality while maintaining low complexity. NYU-Depth v2 results are provided, but the authors' code is not public and we could not reproduce their quality in our implementation and retraining.
\vspace{-4mm}
\paragraph{Convolutional Spatial Propagation Network (CSPN~\cite{cheng_learning_2019})}
was among the first works \cite{cheng_learning_2019, xu_depth_2019, park_non-local_2020} to propose a two-stage network architecture using \emph{spatial propagation} layers. This entails: $(i)$ generation of a ``blurred'' depth map and affinity matrix
; $(ii)$ iterative refinement using the learned affinity matrix as weights for anisotropic diffusion. We retrained the authors' model via their code.
\vspace{-4mm}
\paragraph{Non-Local Spatial Propagation Network (NLSPN~\cite{park_non-local_2020})}
improves upon~\cite{cheng_learning_2019}, its main novelty being the use of non-local neighbors in the diffusion process, which is implemented by deformable convolutions. The authors show it achieves superior performances w.r.t. S2D and CSPN on NYU-Depth v2. In our experiments, we confirm this is indeed the most competitive approach at its network complexity. 
We retrained the authors' model via their code.
\subsection{Neural Networks Quantization}
Quantized Neural Networks (QNN)~\cite{Han15, li2016, zhou2017} are DNNs for which a smaller number of bits $b \ll 32$ is used to represent the weights and 
activations. QNNs are key to achieving efficient and low-power inference: they require considerably less memory and have a lower computational complexity, since quantized values can be stored, multiplied and accumulated efficiently. 

Quantization-Aware Training (QAT) refers to a set of special algorithms to train QNNs, leading to superior results than post-training quantization. Recent QAT methods~\cite{liu2019, cardinaux2018, proxQuant2019, choi2018pact, Jain2020_TQT} proved effective on classification tasks.

These methods often use uniform quantization and the same bit width in each network layer. To improve QAT, Uhlich~\etal~\cite{uhlich_mixed_2020} and Nikoli\'c~\etal~\cite{nikolic2020bitpruning} proposed to \emph{learn} the optimal bit width for each layer while achieving a target size budget. On classification tasks, these \emph{mixed precision} approaches show superior results since the bit widths can be allocated optimally across the network. 
Much less evidence of successful QAT exists for regression tasks such as depth completion. 
\section{Sparse ToF Datasets}
\label{sec:dataset}
\begin{figure}[t]
    \centering
    \setlength{\fboxsep}{0pt}
    \renewcommand{\framebox}{}
    \subfloat[$1216 \times 352, K \approx 4.4 \%$]{
        \framebox{\includegraphics[width=2.6in]{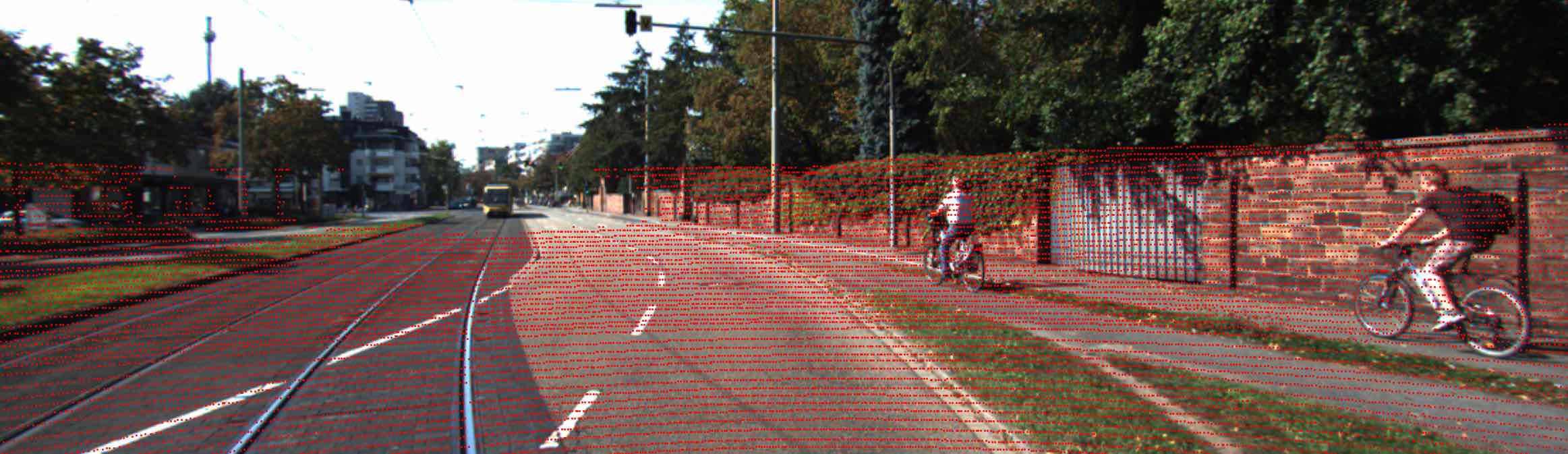}}
    }\\
    \null\hfill
    \subfloat[\label{fig:nyuex}$304 \times 224, K \approx 1.4 \%$]{
        \framebox{\includegraphics[width=1.25in]{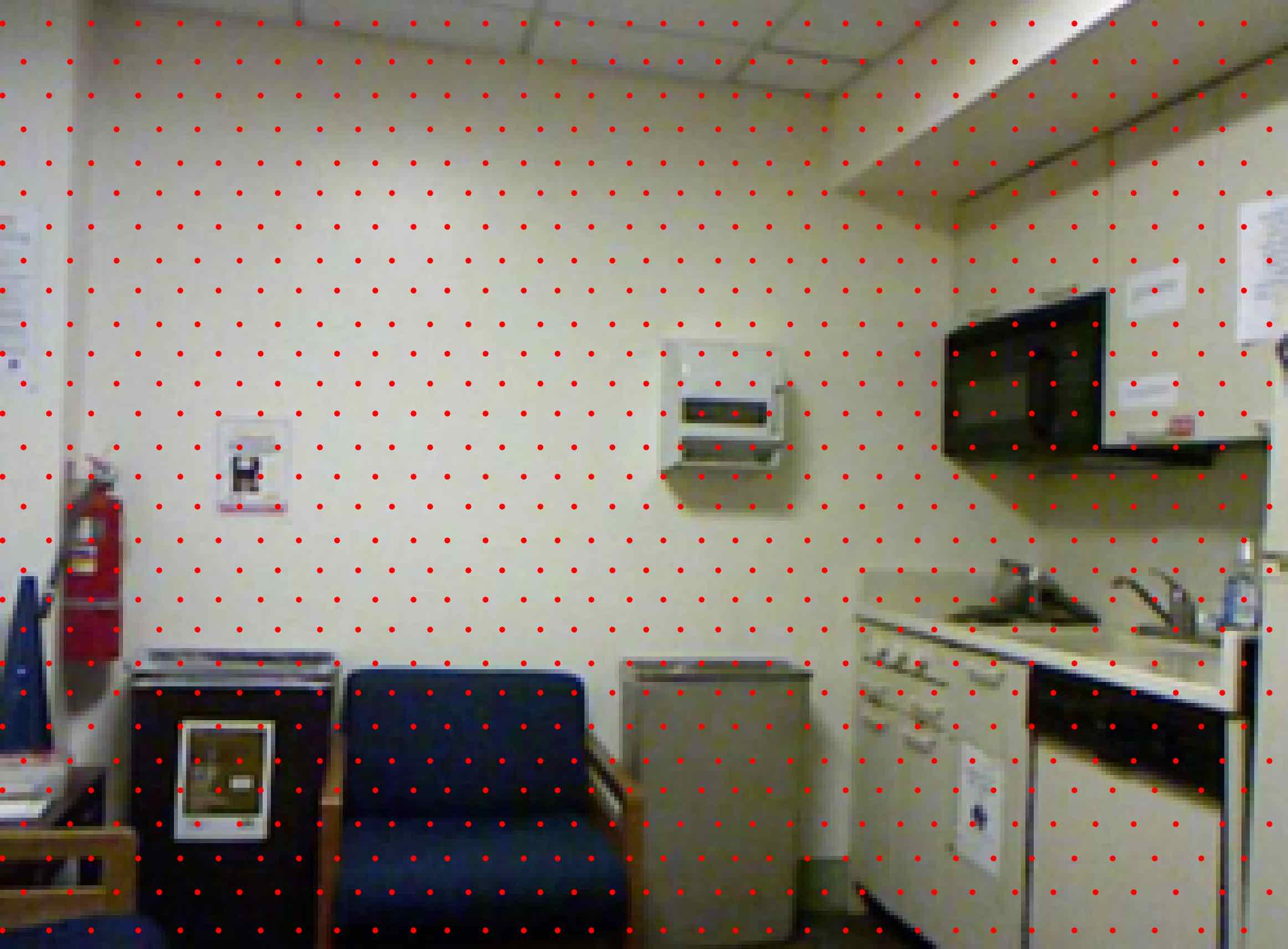}}
        \vspace{0.005in}
    }\hspace{0.025in}
    \subfloat[\label{fig:sdsex}$640 \times 480, K \approx 0.4 \%$]{
        \framebox{\includegraphics[width=1.25in]{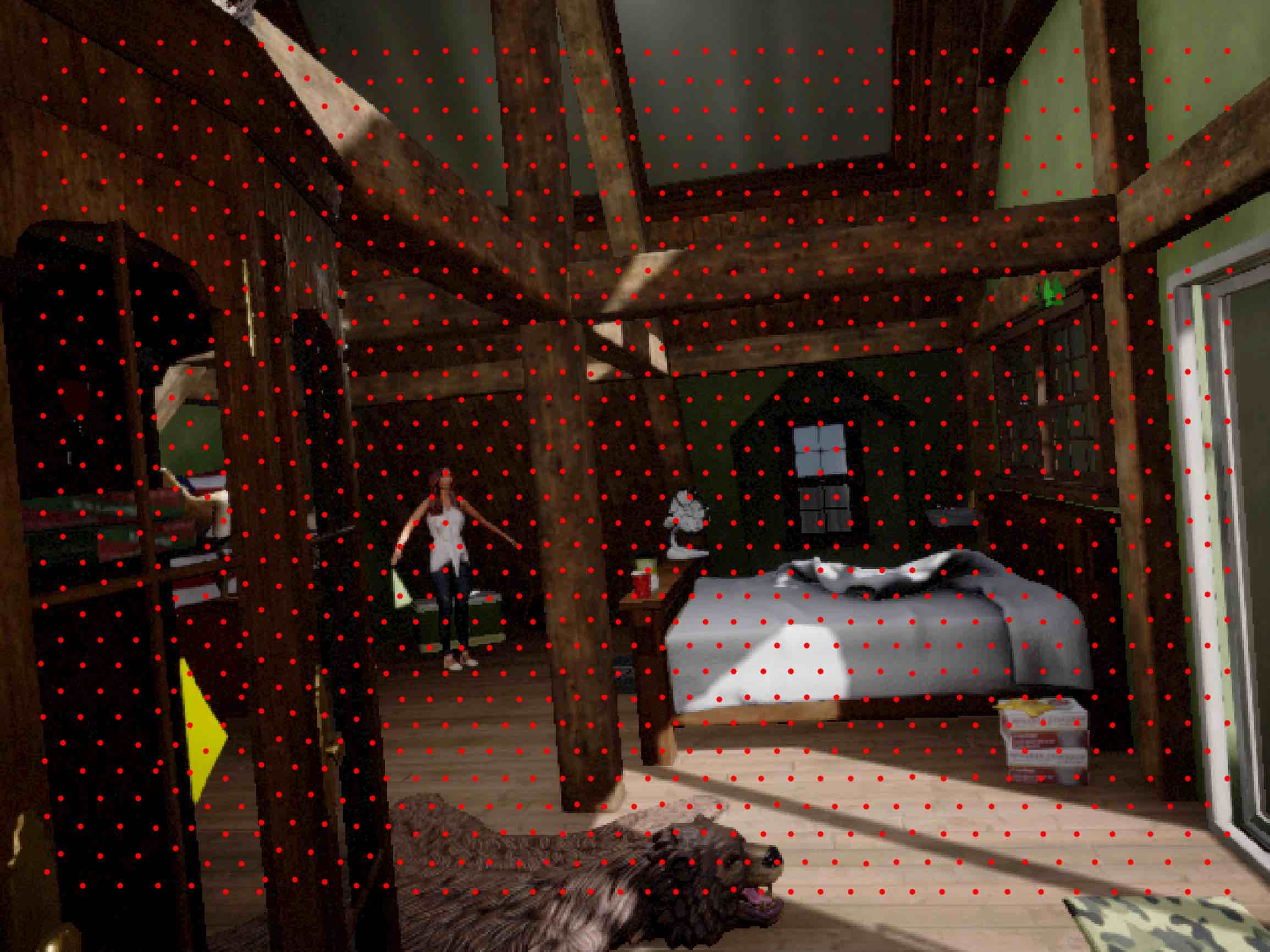}}
    }
    \hfill\null
    \caption{RGB-annotation overlay of (a) KITTI (LiDAR; range: $[0, 85] m$), (b) NYU-Depth v2 (processed; range: $[0, 10] m$), (c) \sdsdata~(sparse ToF; range: $[0, 15] m$). Annotations (red) at available sparse depth coordinates. Figure best viewed in color.
    \label{fig:kittivsdataset}}
    \vspace{-5mm}
\end{figure}

\begin{figure*}[t]
\centering
\subfloat[\label{fig:networkandtraining}]{
\includegraphics[scale=0.4]{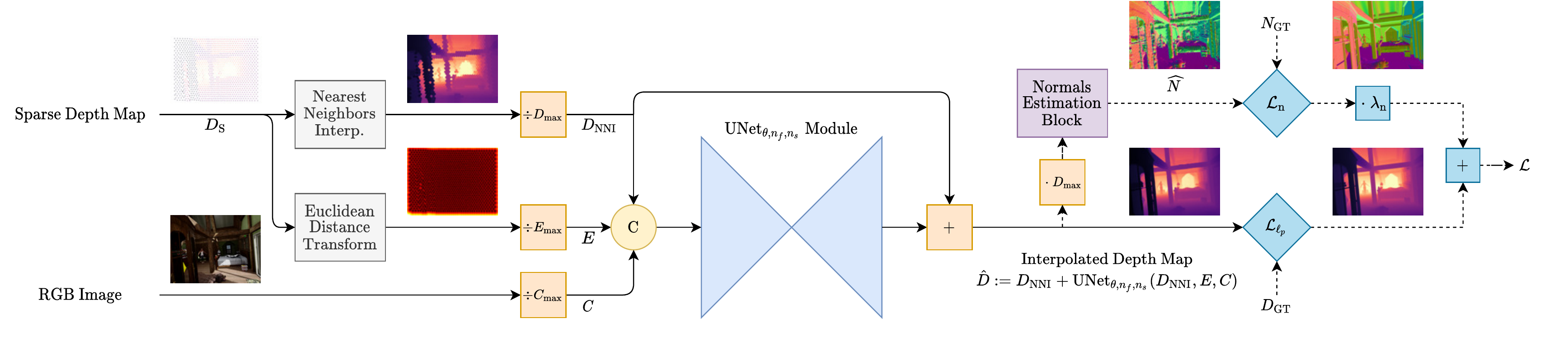}
}
\vspace{0ex}
\null\hfill
\subfloat[\label{fig:unet}]{
\includegraphics[scale=0.15]{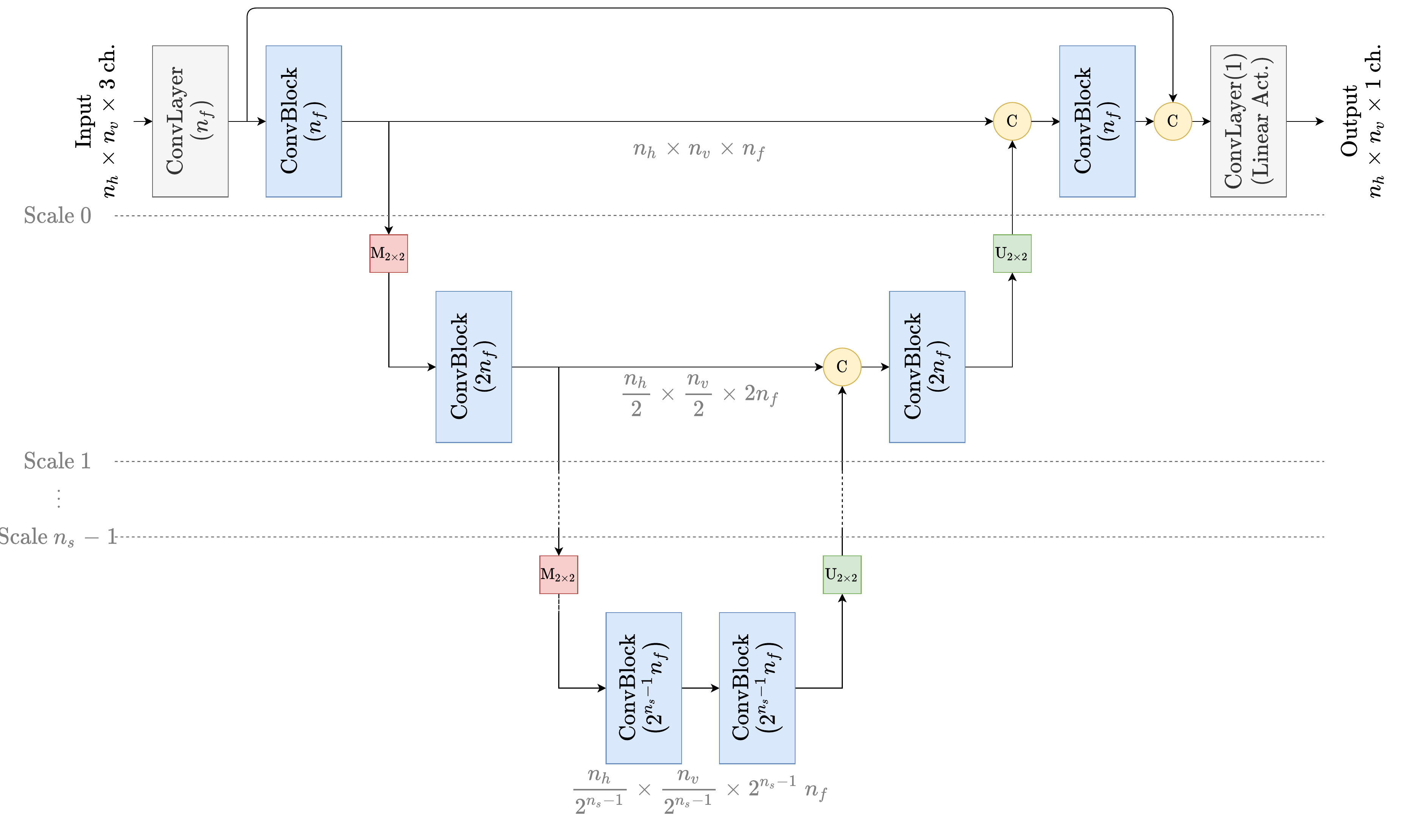}
}
\hfill
\subfloat[\label{fig:convblocks}]{\includegraphics[scale=0.25]{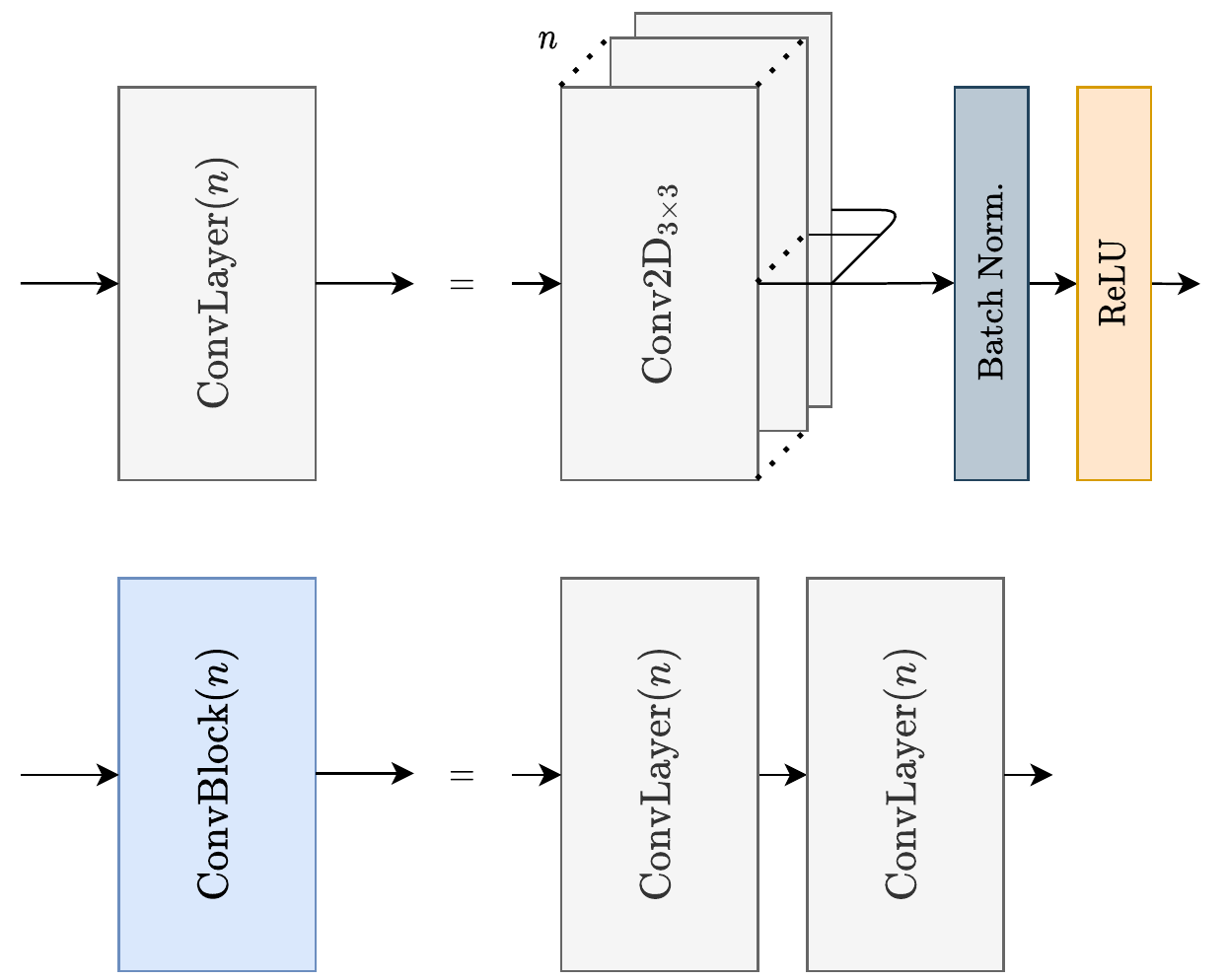}
\vspace{.2in}}
\hfill
\subfloat[\label{fig:normest}]{\includegraphics[scale=0.4]{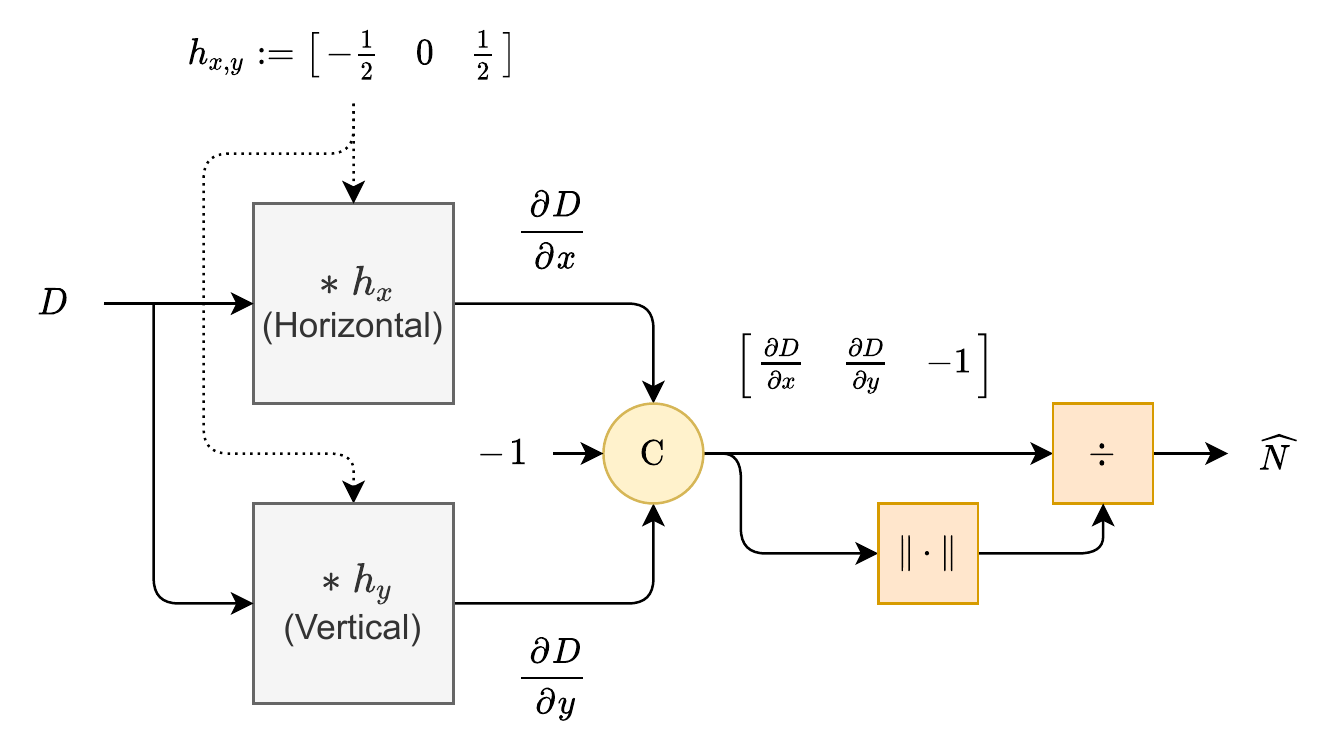}
\vspace{.1in}}
\hfill\null
\caption{Summary of our network design for depth completion: (a) Network architecture. We highlight pre-processing blocks (gray); tensor operators (orange, square); training supervision operators (teal) and connections (dashed); concatenation (yellow, circle); the $\text{UNet}_{\theta, n_f, n_s}$ module (blue); the normals estimation block (purple). (b) $\text{UNet}_{\theta, n_f, n_s}$ Module. $\rm M_{2 \times 2}$ indicates a $\text{MaxPooling2D}$ layer with stride $2$. $\rm U_{2 \times 2}$ indicates an $\text{Upsample2D}$ layer using nearest neighbors upsampling with factor $2$.  (c) Basic convolutional building blocks of the $\text{UNet}_{\theta, n_f, n_s}$ Module. (d) Normals Estimation Block, as implemented by 1-D convolution ($\ast$) in the respective directions.\label{fig:our_method}}
\vspace{-5mm}
\end{figure*}
We target datasets that are critically challenging in terms of sparsity. We consider indoor scenes (range: $[0, 15] m$) with rich structure and detailed objects with diverse camera orientations. Given sparse depth maps $D_{\rm S}$, let us define the \emph{sparsity level} $K := 100  \frac{|\{i \in [n_h] \times [n_v] : {(D_{\rm S}})_i \neq \operatorname{Invalid}\}|}{n_h n_v}$ at target depth map resolution\footnote{Typically equal to the input RGB frame resolution or a cropping of it.} $n_h \times n_v$. 
We tackle very sparse settings where $K \approx 0.4 \sim 1.4\%$ (as a reference in the widely used KITTI depth completion benchmark\footnote{By computing $K$ over the standard validation set of KITTI Depth Completion.} $K \approx 4 \sim 5\%$). A visual overview is provided in~\cref{fig:kittivsdataset}. 

Indeed, such higher values of $K$ simplify the depth completion task. In this work we will focus on indoor 3D perception datasets with very sparse depth maps, where traditional pipelines~\cite{ku_defense_2018} will struggle to achieve good results.

Our main results are given on NYU-Depth v2~\cite{silberman_indoor_2012} as broadly adopted public-domain dataset. For the training set, we use a subset of $\approx 50 \rm K$ RGB-D images from the $249$ standard training scenes. Each image is downsized to $320 \times 240$ then center-cropped to $304\times 224$, identically to~\cite{ma_sparse--dense_2018, park_non-local_2020}. The main difference we introduce is that we process the depth by subsampling it with a triangular tiling dot pattern generated from sparse illumination of commercial VCSELs~\cite{luetzenburg_evaluation_2021} instead of choosing random indices. This results in sparse depth maps $D_{\rm S}$ with on-average $943$ active pixels ($K \approx 1.4\%$; see \cref{fig:nyuex}) sampled from the full, dense depth map $D_{\rm GT}$ which is used as supervision. The surface normals map $N_{\rm GT}$ was also estimated from $D_{\rm GT}$. The standard test set of $654$ images is processed identically.

In addition, we provide results on \sdsdata~\cite{anon_sds-st1k_2021}, a contributed dataset using $8$ diverse environments from Unreal Engine~\cite{unrealengine} marketplace assets yielding $\approx 18\rm K$ images from random camera poses at resolution $640 \times 480$. To obtain sparse ToF data, we apply a light source with dot pattern light shading in our raytracing ToF simulator; this is received on a simulated ToF sensor compatible with low power designs and camera specifications (\eg, sensor integration time, camera intrinsics). This provides both sparse ToF depth maps $D_{\rm s}$ with realistic parallax and pattern geometry, and ground truth $D_{\rm GT}, N_{\rm GT}$. 
In this setting we obtain on-average $1239$ active pixels
, yielding a challenging $K\approx 0.4\%$ (see \cref{fig:sdsex}). 
The recommended training/testing split yields $\approx 5\rm K$ images as test set; this will be provided at \url{https://github.com/sony/ai-research-code}. 
We shall train our models with $160\times160$ non-overlapping patches (\ie, $12$ per image) to meet training device memory limitations. 

\section{Network Design}
The input of all considered depth completion methods is an RGB-D frame comprised of a color image and a sparse depth map, suitably projected to RGB camera space by assuming known intrinsics and RGB-ToF extrinsics, and with negligible RGB-ToF baseline to avoid occlusions. 
%
\begin{figure}
    \centering
    \subfloat[\label{tab:loss-function-sds-vga}]{
    \scriptsize
    \begin{tabular}{cccc}
        \trule
        \rowcolor{gray!10} $\boldsymbol{\mathcal{L}}$ & $\boldsymbol{\lambda_{\rm n}}$ & \bf RMSE ($mm$) &  \bf MNS \\
        \mruleb
        $\mathcal{L}_{\ell_2}$ &                     n/a &    105.2 &  0.538 \\
        $\mathcal{L}_{\ell_1}$ &                    n/a &    {97.8} &  0.754 \\
        $\mathcal{L}_{\ell_1}+\lambda_{\rm n} \mathcal{L}_{\rm n}$ & $1$ &     259.0 &  0.864 \\
        $\mathcal{L}_{\ell_1}+\lambda_{\rm n} \mathcal{L}_{\rm n}$ & $10^{-1}$ &     227.0 &  0.868 \\
        $\mathcal{L}_{\ell_1}+\lambda_{\rm n} \mathcal{L}_{\rm n}$ & $10^{-2}$ &     101.5 &  0.838 \\
        \rowcolor{yellow!30} $\mathcal{L}_{\ell_1}+\lambda_{\rm n} \mathcal{L}_{\rm n}$ & $10^{-3}$ &      98.7 &  0.783 \\
        $\mathcal{L}_{\ell_1}+\lambda_{\rm n} \mathcal{L}_{\rm n}$ & $10^{-4}$ &     109.3 &  0.764 \\
        \brule
    \end{tabular}
    \vspace{-4mm}
    }
    \vspace{-4mm}
    \newcommand{\figSizeAb}{24ex}
    \setlength{\fboxsep}{0pt}
    \subfloat[\label{fig:errors_vs_loss}]{
    \scriptsize
    \begin{tabular}{cc}
    \trule
    \rowcolor{gray!10} \bf Color & $\boldsymbol{\mathcal{L}_{\ell_2}}$ \\
    \specialrule{\arrayrulewidth}{0mm}{2mm}
    \framebox{\includegraphics[width=\figSizeAb]{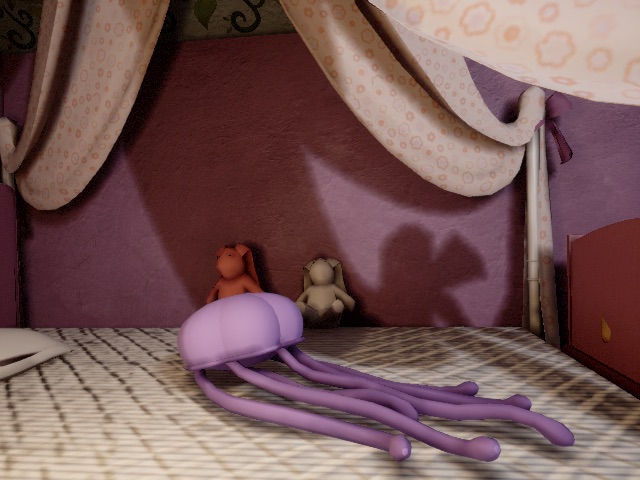}}&
    \framebox{\includegraphics[width=\figSizeAb]{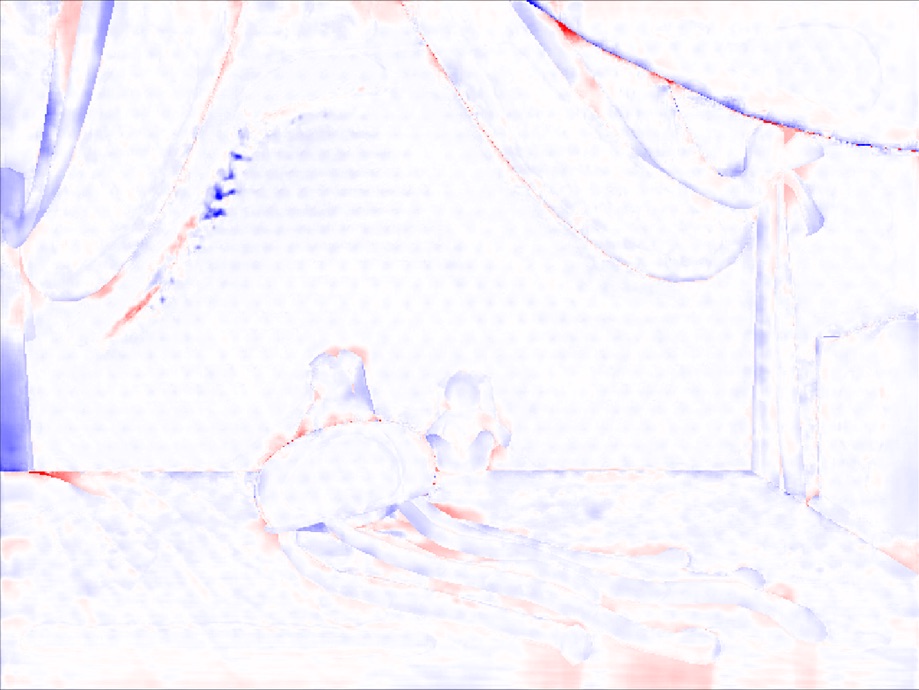}} \\
    \specialrule{\arrayrulewidth}{1mm}{0mm}
    \rowcolor{gray!10}  $\boldsymbol{\mathcal{L}_{\ell_1}}$ & $\boldsymbol{\mathcal{L}_{\ell_1}} + \boldsymbol{{\lambda}^{\rm opt}_{\rm n} \mathcal{L}_{\rm n}}$\\
    \specialrule{\arrayrulewidth}{0mm}{2mm}
    \framebox{\includegraphics[width=\figSizeAb]{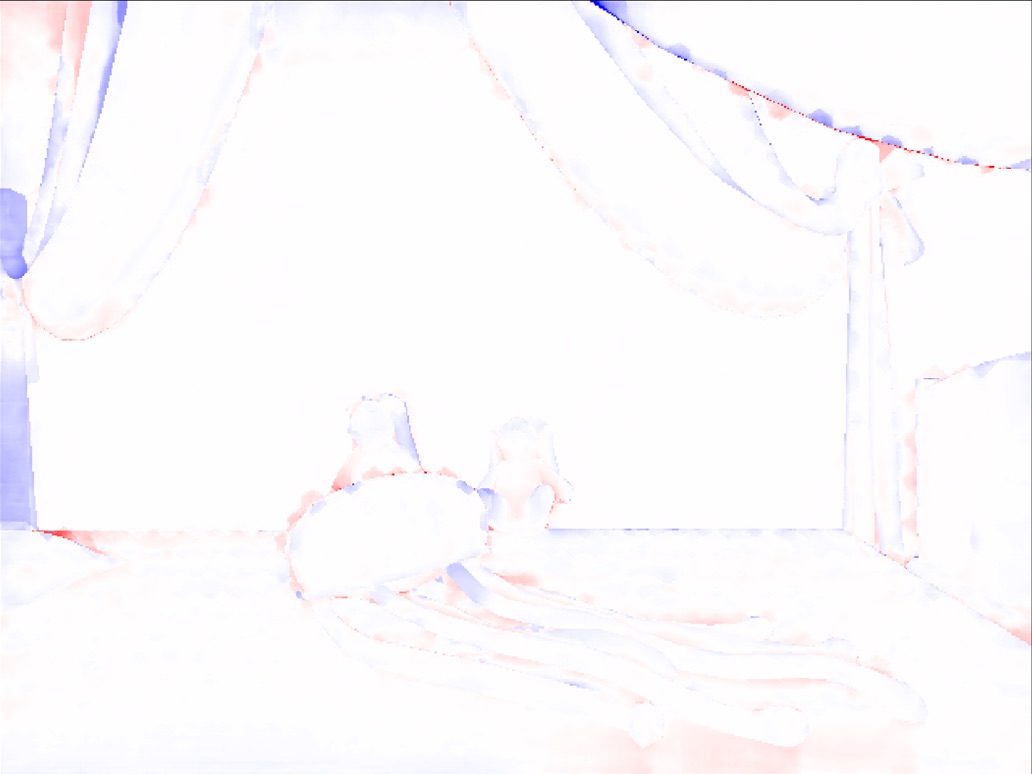}}&
    \framebox{\includegraphics[width=\figSizeAb]{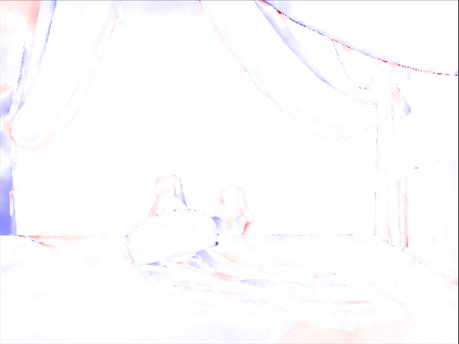}}
    \end{tabular}
    }
    \caption{{Loss function tuning on \sdsdata~(resolution $640 \times 480$): (a) Quantitative analysis; (b) Qualitative analysis of error maps (range: $[-500, 500]\,mm$). Figure best viewed in color.}}
    \vspace{-6mm}
\end{figure}
\subsection{Input Pre-Processing}
We leverage the input pre-processing method of Chen \etal~\cite{chen_estimating_2018}: given the sparse depth $D_{\rm S}$, we compute an initial guess $D_{\rm NNI}$ of the depth map by Nearest Neighbors Interpolation (NNI), \ie, mapping each coordinate to the closest sparse depth pixel in the Euclidean sense. We also compute the Euclidean Distance Transform (EDT) $E$ with~\cite{felzenswalb_distance_2012}~(via OpenCV) given the 2-D valid coordinates of $D_{\rm S}$. NNI is obtained by the same call, which takes median CPU time $t_{\rm NNI} = 4.99~ms$ over \sdsdata~and $t_{\rm NNI} = 0.99~ms$ over NYU-Depth v2. We normalize $D_{\rm NNI}$ by the maximum range $D_{\max}:= 15 m$ and $E$ by $E_{\max} := 40$, which we find empirically to be an appropriate normalization value for our datasets. 
The EDT feeds the CNN an uncertainty map of $D_{\rm NNI}$, \ie, it is low where sparse depth is known and progressively higher when it is not. 
We tried replacing NNI with linear interpolation\footnote{Linear interpolation on a non-uniform grid due to the dot pattern requires Delaunay triangulation, which is costly and cannot be reused as the pattern can change at every frame.} in preliminary tests, but this did not show any benefits while increasing CPU time.
\subsection{Network Architecture}
\label{sec:arch}
A summary of our supervised approach is given in \cref{fig:our_method}. The core CNN element (\cref{fig:unet}) is a simple ``UNet-like''~\cite{ronneberger_u-net_2015} template, dubbed  $\text{UNet}_{\theta, n_f, n_s}$, of which we tune the number of scales $n_{s}$ and feature maps $n_{f}$ relative to the highest scale. Our reference is $n_s = 5, n_f = 64$. The number of feature maps doubles at each scale as their spatial resolution is reduced, up to $2^{n_s - 1}$ at the lowest scale. The basic building blocks (\cref{fig:convblocks}) are standard 2-D convolutions with $3 \times 3$ filters, with padding to same resolution as the layer input and stride $1$. We chose to replace transpose convolutions in the decoder with upsampling and convolution layers, as preliminary tests indicated better performance on our datasets. All skip connections are by concatenation except for the last one, which is a tensor addition w.r.t. $D_{\rm NNI}$. Thus, we have the (normalized) interpolated depth $\hat{D} := D_{\rm NNI} + \text{UNet}_{\theta, n_f, n_s}(D_{\rm NNI}, E, C)$, \ie,  $\text{UNet}_{\theta, n_f, n_s}$ computes a residual w.r.t. the initial guess.
\subsection{Normals Estimation Block}
\label{sec:normals}
As shown in \cref{fig:networkandtraining}, we estimate the surface normals map~$\hat{N}$ from the interpolated depth map~$\hat{D}$ rather than using costly dedicated decoders~\cite{qiu_deeplidar_2019}. We resort to a well-known approximation that is directly applicable to depth maps; the computation graph is reported in~\cref{fig:normest}. Given a depth map tensor $D$ at the input, we approximate horizontal and vertical image axes derivatives by centered differences using two fixed-kernel 1-D convolutions yielding the tensors $(\frac{\partial D}{\partial x},\frac{\partial D}{\partial y})$. We then have at the $i$-th pixel the normal vector
\begin{equation}
(\hat{N})_i := \textstyle\frac{\begin{bmatrix}(\frac{\partial D}{\partial x})_i & (\frac{\partial D}{\partial y})_i & -1\end{bmatrix}}{\sqrt{(\frac{\partial D}{\partial x})^2_i + (\frac{\partial D}{\partial y})^2_i + 1}}.
\end{equation}
We found this small-kernel method sufficiently accurate to enforce similarity\footnote{We ensure the surface normals' orientation is always consistently pointing towards the camera both in $N_{\rm GT}$ and $\hat{N}$.} between $(\hat{N})_i, (N_{\rm GT})_i$ at training.
\subsection{Loss Function}
\label{sec:loss}
Our supervised training follows the flow in \cref{fig:networkandtraining}. The supervision is comprised of two terms. The first term measures the $\ell_p$-norm distance between (normalized) ground truth $D_{\rm GT}$ and $\hat{D}$, \ie, for dataset $\mathcal{D}$ with $J$ samples, \begin{equation}
\textstyle\mathcal{L}_{\ell_p}(\mathcal{D}) := \frac{1}{J}\sum^{J}_{j = 1}\|\operatorname{vec}(\hat{D_j} \!-\! D_{{\rm GT}, j})\|^p_p, \ p = 1, 2.
\end{equation}
The second term measures similarity between the dense ground truth normals map $N_{\rm GT}$ and~$\hat{N}$ from the block in \cref{fig:normest}: for each $i$-th pixel normal we measure the cosine similarity $\cos \alpha = \langle (N_{{\rm GT}})_i, (\hat{N})_i \rangle$. We then take its average over the dataset and 
define the normals loss \begin{equation}
\textstyle\mathcal{L}_{\rm n}(\mathcal{D}) := -\,\frac{1}{J n_h n_v}\sum^{J}_{j = 1} \sum^{n_h n_v}_{i = 1} \langle (N_{{\rm GT}, j})_i, (\hat{N_j})_i\rangle
\end{equation}
with range $[-1, 1]$ (at $-1$ all normals are identical). 
In the presence of invalid ground truth depth or normals map (\eg, for invalid pixels in NYU-Depth v2) we exclude the corresponding pixel from the loss computations.

Only part of the literature uses normals supervision\footnote{Smoothness losses on the gradients of $\hat{D}$ are also common \cite{ma_self-supervised_2019, shivakumar_dfusenet_2019} but not considered here as the normals loss implements similar behavior in a supervised and more geometrically sound fashion.}~\cite{qiu_deeplidar_2019, xu_depth_2019}, while all depth completion approaches use depth supervision providing scale. Normals supervision alone is not sufficient to retrieve scale. Thus, we minimize
\begin{equation}
\textstyle\mathcal{L}(\mathcal{D}) := \mathcal{L}_{\ell_p}(\mathcal{D})+\lambda_{\rm n}\mathcal{L}_{\rm n}(\mathcal{D})\label{eq:theloss}
\end{equation}
\noindent which balances between scale-dependent and scale-independent terms using $\lambda_{\rm n} \in [0, 1]$. This weight depends on the dataset and must be tuned empirically. 
To do so, we run preliminary tests of our approach~(\cref{fig:networkandtraining}) on \sdsdata~under different\footnote{We verified in preliminary tests that $\mathcal{L}_{\ell_2} + \lambda_{\rm n} \mathcal{L}_{\rm n}$ is systematically worse than the $\ell_1$ case and excluded it accordingly.} $\mathcal{L} := \{\mathcal{L}_{\ell_2}, \mathcal{L}_{\ell_1}, \mathcal{L}_{\ell_1}+\lambda_{\rm n} \mathcal{L}_{\rm n}\}$ and $\lambda_{\rm n} := 10^{-q}, q = 0, \ldots, 4$, in the same settings described in~\cref{sec:setupdc}. Indeed, defining the conventional RMSE and Mean Normals Similarity
\begin{equation}
\textstyle\operatorname{MNS}:= \frac{1}{J n_h n_v}\sum^{J}_{j = 1} \sum^{n_h n_v}_{i = 1} \langle (N_{{\rm GT}, j})_i, (\hat{N}_{j})_i\rangle\label{eq:MNS}
\end{equation}
\noindent as quality metrics we see that large values of $ \lambda_{\rm n}$ degrade the RMSE, and \emph{vice versa} for MNS. We find the best trade-off at ${\lambda}^{\rm opt}_{\rm n} = 10^{-3}$. As confirmation of this, in \cref{fig:errors_vs_loss} we report one sample error map $\hat{D} - D_{\rm GT}$ under different $\mathcal{L}$. As we move from $\mathcal{L}_{\ell_2}$ to $\mathcal{L}_{\ell_1}$ the observed error map pattern artefact disappears. Remaining errors at depth map discontinuities are reduced by using the normals loss $\mathcal{L}_{\rm n}$.

\subsection{Network Tuning}
\label{sec:hyper}
\begin{table}[t]
    \begin{center}
    \scriptsize
    \begin{tabular}{ccccc}
    \trule
    \rowcolor{gray!10}$\boldsymbol{n_f}$ & $\boldsymbol{n_s}$ &  \bf MParams &   \bf RMSE ($mm$) &  \bf MNS \\
    \mruleb
    64 &         5 &    50.3 &       98.7 &      0.783 \\
    32 &         5 &    12.6 &      105.0 &      0.775 \\
    16 &         5 &     3.2 &      181.1 &      0.751 \\
    64 &         4 &    12.6 &      219.2 &      0.779 \\
    64 &         3 &     3.1 &      248.4 &      0.761 \\
    32 &         4 &     3.1 &      274.8 &      0.767 \\
    \bottomrule
    \end{tabular}
    \end{center}
    \caption{Network tuning on \sdsdata~(resolution $640 \times 480$). We confirm a steady decrease of quality metrics as the feature maps $n_f$ relative to the highest scale and the scales $n_s$ are reduced.}
    \label{tab:a-asds-vga}
    \vspace{-6mm}
\end{table}
We investigated simple strategies to reduce the size of the initial model, such as tuning the $\text{UNet}_{\theta, n_f, n_s}$ module by its number of feature maps $n_{f}$ relative to the highest scale, and the number of scales $n_{s}$.  Our preliminary tests, in the same setting as \cref{sec:setupdc} on \sdsdata, actually confirm  (\cref{tab:a-asds-vga}) that any pair smaller than $(n_f, n_s) = (64, 5)$ leads to a rapid degradation of RMSE and MNS, with the reduction of $n_s$ (shallower network) being generally worse than that of $n_f$. Thus, we choose to achieve memory size reduction by quantization of the full model with $(n_f, n_s) = (64, 5)$ which, as we will show, allows for a more compact model with limited impact on quality metrics. 
\section{Quantization-Aware Training}
\label{sec:quantization}
Let $Q(x; \boldsymbol{\phi})$ be an arbitrary scalar quantizer with parameters $\boldsymbol{\phi}$, which maps
$x \in \mathbb{R}$ to discrete values $\{ q_1, ..., q_I\}$ represented by 
$b$ bits, $b: I \leq 2^b$. This quantizer may be applied to both weights and activations to reduce their memory size requirements. This can be done by either post-training quantization or by Quantization-Aware Training (QAT), which generally achieves better accuracy than the former. QAT consists in training the quantized version of a DNN (in this case, CNN) while applying $Q(x; \boldsymbol{\phi})$ to its weights and/or activations. However, this is challenging as the gradient through the quantizer is zero almost everywhere; a solution to this is the Straight-Through Estimator (STE)~\cite{bengio2013estimating}. 
A \emph{symmetric uniform} quantizer $Q_U(x; \boldsymbol{\phi})$ then maps $x \in \mathbb{R}$ to 
uniformly quantized values by
\begin{equation}
    q = Q_U(x; \boldsymbol{\phi}) := \text{sign}(x) \textstyle\begin{cases}
		  d \floor{\frac{\lvert x \rvert}{d}+\frac{1}{2}} & \lvert x \rvert \le \xm \\
		  \xm & \lvert x \rvert > \xm
    \end{cases}, \label{eq:uniform_quantization}
\end{equation}
\noindent where the parameter vector $\boldsymbol{\phi}:=[d, \xm, b]^T$ with step size $d \in \R^+$, maximum value $\xm \in \R^+$, and number of bits or \emph{bit width} $b \in \mathbb{N}: b \ge 2$ of the quantized value $q$. 
\subsection{Uniform Precision Models}
We start by considering \emph{uniform precision} (UP) QAT, \ie, every weight (or activation) of every layer has the same predefined bit width $b$ and QAT optimizes over the remaining parameters $d, \xm$ of $\boldsymbol{\phi}$. 
We conducted extensive experiments to preserve depth completion accuracy while quantizing with \eqref{eq:uniform_quantization} the weights and activations in our model. The most effective UP QAT training procedure was obtained by initializing from the pretrained $\rm float32$ model and using learnable $\xm$ as in the Trained Quantization Threshold (TQT) approach~\cite{Jain2020_TQT}; to stabilize the QAT, we used  cosine learning rate decay scheduling~\cite{loshchilov2017}, and we first trained with RMSprop until convergence ($32$ epochs in our case), followed by Adam~\cite{kingma2014adam} for $20$ more epochs.

\subsection{Mixed Precision Models}
Recent results~\cite{uhlich_mixed_2020, nikolic2020bitpruning} show that, for a given memory size budget, better accuracy than UP QAT can be achieved using \emph{mixed precision} (MP) QAT, \ie, each layer uses its own bit width \emph{learned at training} to fit a target total network size. Following~\cite{uhlich_mixed_2020} we parametrize $Q_U(x; \boldsymbol{\phi})$ with their range and step size $d$, from which the bit width $b :=  \lceil 1 + \log_2 (\xm/d + 1)\rceil$ 
is then inferred. With this parametrization and STE~\cite{bengio2013estimating} gradient, we can use standard stochastic gradient descent methods to learn  per-layer optimal bit widths jointly with network parameters.
To achieve target network sizes we add to \eqref{eq:theloss} \emph{size constraints} on the weights and activations as penalty terms~\cite{bertsekas2014constrained}, minimizing
\begin{align}
    & \textstyle\mathcal{L}_{\rm MP} = \mathcal{L}(\mathcal{D}) + \lambda_{\rm W} \max(0, S_{\rm W})^2 + \lambda_{\rm A}\max(0, S_{\rm A})^2 \label{eq:penalty_opt_prob} \\
    & S_{x} = \textstyle(\sum^{L}_{l=1} S^l_{x}) - S^o_{x}, S^l_{x} = N_{{x}, l} b_{{x}, l}, S^o_{x} = \bar{b}_{{x}} \sum^{L}_{l=1} N_{{x}, l} \label{eq:sizes}
\end{align}
where $x := \{\rm W, \rm A\}$ denote weights or activations; $S^l_{x}$ their per-layer total memory size for $N_{{x}, l}$ coefficients with learned bit widths $b_{x, l}$; $S^o_{x}$ is the \emph{target size} which we tune\footnote{This tuning can arbitrarily refer to an average bit width or to a total size (in which case the average as defined in \eqref{eq:sizes} can be fractional).} by a reference \emph{average bit width} $\bar{b}_{{x}}$. 
The parameters $\lambda_{\rm W} \in \mathbb{R}^+$ and $\lambda_{\rm A} \in \mathbb{R}^+$ are chosen to balance the respective penalties. We empirically set them following the criteria in \cite[Sec. 4]{uhlich_mixed_2020}, yielding
$\lambda_{\rm W} \approx 2.66 \cdot 10^{-7}, \lambda_{\rm A} \approx 1.73 \cdot 10^{-6}$. 
In MP QAT we also initialize network parameters from the pretrained $\rm float32$ model, and run Adam for $60$ epochs until convergence with cosine learning rate decay scheduling.
\subsection{Weights Quantization}
\label{sec:weightsq}
\begin{figure}
    \centering
    \subfloat[\label{tab:quantization_unif}]{
    \centering
    \scriptsize
    \begin{tabular}{cccc}
    \trule
    \rowcolor{gray!10} \bf Precision & \bf Weights Size (MB) & \bf RMSE ($mm$) &  \bf MNS \\
    \mruleu
    \rowcolor{gray!5} $\text{float32}$ & 198.5 & 98.7 & 0.783 \\
    Uniform ($\rm W^{\rm UP}_4$) & 25.1 & 100.3 & 0.762 \\
    Mixed ($\rm W_{\star}$) & \bf 13.9 & \bf 94.5 & \bf 0.783 \\
    \brule
    \end{tabular}
    \vspace{-4mm}
    }\\
    \subfloat[\label{fig:bw-per-layer}]{
    \centering
    \begin{tikzpicture}
    \begin{axis}[
        width=3.2in,
        height=1.2in,
        cycle list/Dark2,
        xmin=-0.5,
        xmax=28.5,
        xlabel={Layer Index $l$},
        xlabel style={
        at = {(0.5,-0.15)},
        font=\scriptsize
        },
        ytick={0,2,4,6,8,10,12},
        xtick align=inside,
        ymin=0,
        ymax=12.5,
        ylabel={Bit Width $b_{{\rm W}, l}$},
        ylabel style={
        at = {(-0.05,0.5)},
        font=\scriptsize
        },
        ymajorgrids,
        axis background/.style={fill=white},
        every tick label/.append style={font=\tiny},
        axis background/.style={fill=white},
        legend style={
        legend columns=-1,
        row sep=0.5pt,
        legend cell align=left,
        align=left,
        fill=white, fill opacity=0.65,
        text opacity=1,
        draw=none,
        font=\fontsize{4}{1}\selectfont
        },
        clip=false
        ]
    \addplot [draw=blue, fill=blue!10, bar width=3pt, ybar] coordinates {
    (0, 12)
    (1, 8)
    (2, 7)
    (3, 6)
    (4, 4)
    (5, 7)
    (6, 3)
    (7, 3)
    (8, 6)
    (9, 3)
    (10, 3) 
    (11, 3) 
    (12, 2) 
    (13, 2) 
    (14, 3)
    (15, 2) 
    (16, 2) 
    (17, 2) 
    (18, 2) 
    (19, 2)
    (20, 3)
    (21, 5)
    (22, 3)
    (23, 4)
    (24, 6)
    (25, 6)
    (26, 8)
    (27, 10)
    (28, 8)
    };
    
    \addplot [draw=violet, thick, densely dashed] coordinates { (-0.5, 2.35) (28.5, 2.35) } node [pos=1, font=\tiny, anchor=west] {\color{violet} $\bar{b}_{\rm W}$};
    \end{axis}
\end{tikzpicture}
    }
    \caption{Weights quantization. (a) Results on \sdsdata~(resolution $640 \times 480$) for the most compact UP model $W^{\rm UP}_4$ at bit width $b = 4$ and MP model $W_{\star}$ at average bit width $\bar{b}_{\rm W} = 2.35$. (b) Per-layer precision allocation of $\rm W_{\star}$ (last $\rm float32$ layer omitted) and the corresponding $\bar{b}_{\rm W}$ (dashed).}
    \vspace{-6mm}
\end{figure}
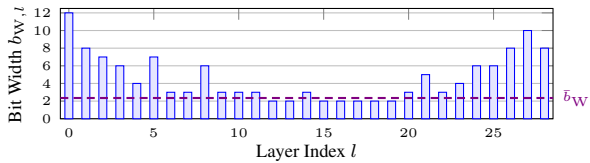
We initially compare \emph{weights-only} UP QAT and MP QAT against the pretrained ${\rm float32}$ model, with the same dataset and setting as \cref{sec:hyper}. 
In all our models the last layer is at $\rm float32$ precision as it directly produces the regression output in \cref{fig:networkandtraining}.
We refer to UP models ${\rm W}^{\rm UP}_{b}$ and MP models ${\rm W}_{b}$ at (fixed or average) bit width $b$. We report the most compact UP model $\rm W^{\rm UP}_4$ ($b=4$) providing best quality metrics, and compare it against the most compact MP model ${\rm W}_{\star}$, where $\star$ denotes target size $S^o_{\rm W, \star} := 14 \, \rm MB$, \ie, $\bar{b}_{\rm W}\approx 2.35 \, \rm bit$ when setting the size constraint.

The results in \cref{tab:quantization_unif} allow us to establish that MP QAT is superior to UP, as with half of the weights memory size we see no degradation of MNS and a slight improvement in RMSE due to refinement from the pretrained $\rm float32$ model. We also observe in \cref{fig:bw-per-layer} how the per-layer precision allocation of MP naturally tends to assign smaller (larger) bit widths to inner (outer) layers of the QNN.
%
\begin{table*}
    \centering
    \scriptsize
    \begin{tabular}{cMMMMMMSSSL}
    \trule
    \rowcolor{gray!10}\bf Method  & \bf Weights Size (MB) & {\bf Activations Size (MB)} & \bf RMSE ($mm$) & \bf  MAE  ($mm$) &  \bf MRE ($\%$) & \bf MNS & $\boldsymbol\delta_1 (\%)$ & $\boldsymbol\delta_2(\%)$ & $\boldsymbol\delta_3(\%)$ & \bf Median $t_{\rm GPU}$ ($ms$) \\
    \mruleu
    \rowcolor{green!10}\multicolumn{11}{c}{\bf NYU-Depth v2 (center cropped, $304 \times 224$)}\\
    \mruleb
        S2D & 90.1 & 322.7 & 72.1 & 34.1 & 1.230 & 0.538 &   99.79 & 99.97 & 99.99 & {\bf 4.83} (0) \\
    $\rm D^3$ & 7.8 & 59.9 & 128.3 & 46.9 & 1.760 & 0.422 &   99.13 & 99.79 & 99.94 & 13.78 (0.99{\tiny $^{\rm CPU}$}) \\
    CSPN & 832.1 &  44.2 & 79.0 &     28.5 &   0.970 &  0.752 &   99.68 &   99.95 &   99.99 &  63.97 (14.02)  \\
    NLSPN & 100.1 & 200.2 & \bf 61.4 & \bf 22.5 & \bf 0.750 &  \bf 0.800 &  \bf 99.83 & \bf 99.98 & \bf 100.00 &  42.50 (7.93) \\
    Ours &  198.5 & 145.5 & \underline{63.7} &     \underline{23.2} &   \underline{0.782} &  \underline{0.790} &   \underline{99.81} &   \underline{99.98} &  \underline{99.99} & \underline{5.11} (0.99{\tiny $^{\rm CPU}$}) \\
    \mruleu
    \rowcolor{blue!10}\multicolumn{11}{c}{\bf \sdsdata~($640 \times 480$)}\\
    \mruleb
        S2D &  90.1 & 1455.5 & 643.3 &    589.9 &  29.130 &  0.158 &   55.45 &   83.27 &   94.25 & {\bf 6.02} (0)\\
    $\rm D^3$ & 7.8 & 270.2 &     176.7 &     57.4 &   2.500 &  0.572 &   98.40 &   99.40 &   99.70 & 18.83 (4.99{\tiny $^{\rm CPU}$}) \\
    CSPN  & 832.1 & 199.2 &     174.3 &     97.1 &   3.850 &  0.687 &   98.60 &   99.60 &   99.78 & 154.62 (23.50) \\
    NLSPN & 100.1 & 890.6 &      \underline{99.1} &     \underline{29.4} &   \underline{1.150} &  \underline{0.775} &   \underline{99.39} & \bf 99.81 &  \bf 99.92 & 148.68 (11.18) \\
    Ours & 198.5 &                656.3 &     \bf 98.7 &     \bf 21.7 &  \bf 0.829 & \bf 0.783 &  \bf 99.45 &   \underline{99.81} &   \underline{99.91} &  \underline{10.30} (4.99{\tiny $^{\rm CPU}$}) \\
    \brule
    \end{tabular}
    \caption{Depth completion models. The median $t_{\rm GPU}$ is given as $t_{\rm total} (t_{\rm initialization})$. In ours and $\rm D^3$, initialization is computed on $^{\rm CPU}$.}
    \label{tab:results-dc}
    \vspace{-5mm}
\end{table*}

\begin{table*}
    \centering
    \scriptsize
    \begin{tabular}{cMMMMMMSSS}
    \trule
    \rowcolor{gray!10} \bf Precision & \bf Weights Size (MB) & \bf Activations Size (MB) & \bf RMSE ($mm$) & \bf  MAE  ($mm$) &  \bf MRE ($\%$) & \bf MNS & $\boldsymbol\delta_1 (\%)$ & $\boldsymbol\delta_2(\%)$ & $\boldsymbol\delta_3(\%)$ \\
    \mruleu
    \rowcolor{green!10}\multicolumn{10}{c}{\bf NYU-Depth v2 (center cropped, $304 \times 224$)}\\
    \mruleb
    \rowcolor{gray!5} $\text{float32}$ &  198.5 & 145.5 & 63.7 &     23.2 &   0.782 &  0.790 &   99.81 &   99.98 &   99.99 \\
    $\rm W_8 A_8$ & 22.1 &  47.9 & \bf 64.2 &     \bf 23.4 &   \bf 0.784 &  \bf 0.793 &  \bf 99.81 &   \bf 99.98 &   \bf 99.99 \\
    \rowcolor{yellow!30} $\rm W_4 A_8$ & \underline{22.0} & \underline{36.2} & \underline{64.3} &    \underline{23.5} &  \underline{0.793} &  \underline{0.790} & \underline{99.81} & \underline{99.98} & \underline{99.99} \\
     $\rm W_4 A_4$ & 23.7 & \bf 17.6 & 70.3 &     26.1 &   0.894 &  0.739 &   99.76 &   99.97 &   99.95 \\
    \rowcolor{orange!30} $\rm W_\star$ &  \bf 13.7 & 145.5 & 64.9 &     23.6 &   0.796 &  0.786 &   99.81 &   99.98 &   99.99\\
    \mruleu
    \rowcolor{blue!10}\multicolumn{10}{c}{\bf \sdsdata~($640 \times 480$)}\\
    \mruleb
    \rowcolor{gray!5} $\text{float32}$ &      198.5 &                656.3 &      98.7 &     21.7 &   0.829 &  0.783 &   99.45 &   99.81 &   99.91\\
     $\rm W_8 A_8$ &       45.9 &                164.1 &     101.7 &     22.5 &   0.874 &  \underline{0.781} &   99.40 &   99.79 &   99.91\\
    \rowcolor{yellow!30} $\rm W_4 A_8$ &       \underline{20.7} &                \underline{163.3} &     101.3 &     \underline{21.1} &   \underline{0.810} &  0.776 &   \underline{99.48} & \underline{99.83} & \underline{99.92}\\
     $\rm W_4 A_4$ &       21.1 &                \bf 78.4 &   \underline{99.0} &     24.4 &   0.958 &  0.638 &   99.41 &   99.81 &   99.92\\
    \rowcolor{orange!30}  $\rm W_\star$ &       \bf 13.9 &                656.3 &      \bf 94.5 &     \bf 20.8 &   \bf 0.808 &  \bf 0.784 &   \bf 99.48 & \bf 99.83 & \bf 99.92\\
    \brule
    \end{tabular}
    \caption{Quantized network models. Among MP QNN models, we highlight the best trade-off between lowest memory footprint and best quality metrics for weights and activations (yellow) and weights-only (orange). Figure best viewed in color.\label{tab:results-mp}}
    \vspace{-3.5mm}
\end{table*}
%
\subsection{Activations Quantization}
\label{sec:activations}
For efficient inference on mobile or low-power devices, it is crucial to consider the quantization of both weights and activations. Indeed, activations can dominate memory requirements at inference (especially for encoder-decoder networks such as ours) and may additionally cost many read/write accesses of buffer memory, with large impact on energy consumption. 
As we observed that MP is capable of achieving superior performances than UP for the weights-only case, our final results (\cref{sec:perf_quantization}) will consider MP QAT with weights and activations quantization at $4$ to $8$ bits; we therefore denote them by ${\rm W}_{b'} {\rm A}_{b''}$ where $b', b'' \geq 2$ are average bit widths that set the respective size constraints. 
\section{Results}
We present our results using the same quality metrics of most prior works~\cite{ma_sparse--dense_2018, chen_estimating_2018}: RMSE, Mean Absolute Error (MAE), Mean Absolute Relative Error (MRE), and $\delta_i, i = 1, 2, 3$; we also add the MNS metric~\eqref{eq:MNS} to study normals similarity. Where reported, the median GPU time per prediction ($t_{\rm GPU}$) is measured on an NVidia GTX 1080 Ti. The reported results are given on full-frame test sets from \cref{sec:dataset}. %

\subsection{Training Setup}
\label{sec:setupdc}
We train our depth completion model for $40$ epochs until convergence using RMSprop as the optimizer with learning rate $\rho := 10^{-4}$ and batch size of $8$ (\sdsdata) or $4$ (NYU-Depth v2). Given its very large resolution, all trainings of our model on \sdsdata{} leverage patches of $160 \times 160$, but we switch to full $640 \times 480$ inference when computing quality metrics since our networks are fully convolutional. 

We also retrain other competing methods for comparison, by following the supervised training procedures described by the respective authors as reproducible from their codes~(S2D, CSPN and NLSPN) or descriptions ($\rm D^{3}$), while providing as input our sparse ToF datasets. We also tested the authors' CSPN and NLSPN models pretrained on NYU-Depth v2 with randomly sampled patterns, but the resulting performances were worse than our retraining.

In addition to the $\rm float32$ model, we consider our quantized MP models $\rm W_{8} \rm A_{8}, \rm W_{4} \rm A_{8}, \rm W_{4} \rm A_{4}$ (\cref{sec:activations}) and $\rm W_\star$ (\cref{sec:weightsq})\footnote{In $\rm W_\star$ activations are left $\rm float32$.}; the subscripts denote the respective average bit widths. For MP QAT, we use the NNabla~\cite{hayakawa_neural_2021} implementation of~\cite{uhlich_mixed_2020}. In this case we obtained the best results by training our MP models starting from the pretrained $\rm float32$ model, with the Adam optimizer for $60$ epochs until convergence, starting from learning rate $\rho := 10^{-4}$ with cosine learning rate decay scheduling~\cite{loshchilov2017}. 

\subsection{Depth Completion Models}
\begin{figure*}[t]
\newcommand{\figSize}{17ex}
\setlength{\fboxsep}{0pt}
\tabcolsep=0.1cm
\centering
\scriptsize
\begin{tabular}{cccccccc}
\trule
\rowcolor{gray!10} & \multicolumn{2}{c}{\bf NLSPN} &  \multicolumn{2}{c}{ \bf Ours ($\text{float32}$) } &  \multicolumn{2}{c}{ \bf Ours ($\rm W_4 A_8$) } & \\
\rowcolor{gray!10} \multirow{-2}{*}{\bf Color}  & \bf Depth Map & \bf Error Map & \bf Depth Map & \bf Error Map & \bf Depth Map & \bf Error Map & \multirow{-2}{*}{$D_{\rm GT}$}\\
\mruleu
\rowcolor{green!10}\multicolumn{8}{c}{\bf NYU-Depth v2 (center cropped, $304 \times 224$)}\\
\specialrule{\arrayrulewidth}{0mm}{2mm}
\framebox{\includegraphics[width=\figSize]{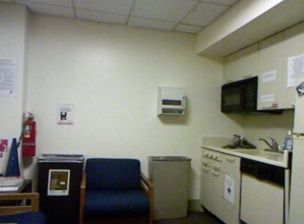}} &
\framebox{\includegraphics[width=\figSize]{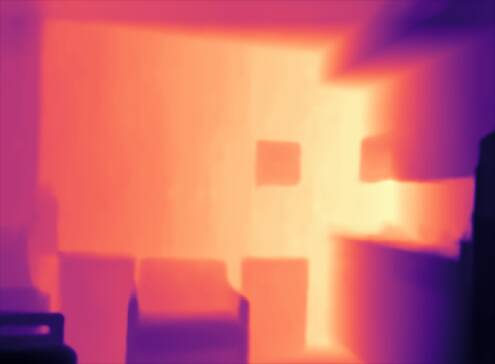}} &
\framebox{\includegraphics[width=\figSize]{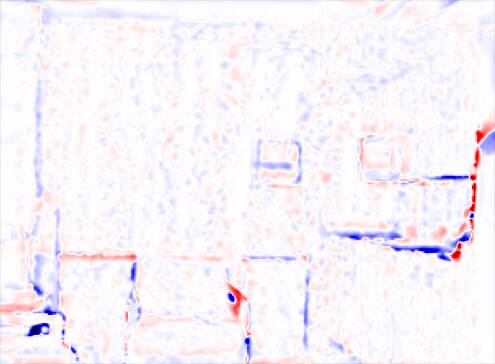}} &
\framebox{\includegraphics[width=\figSize]{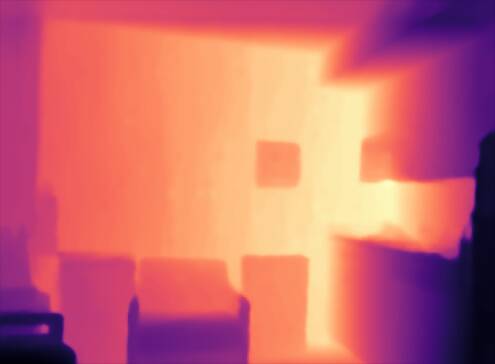}} &
\framebox{\includegraphics[width=\figSize]{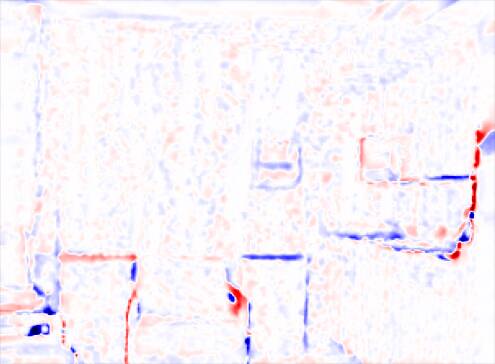}} &
\framebox{\includegraphics[width=\figSize]{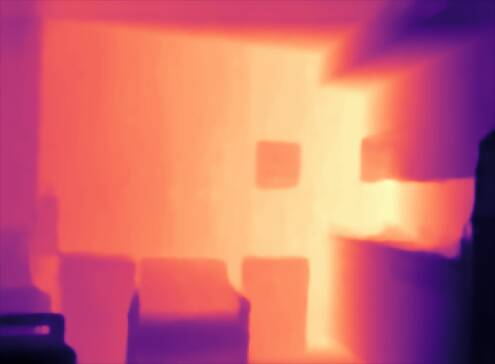}} &
\framebox{\includegraphics[width=\figSize]{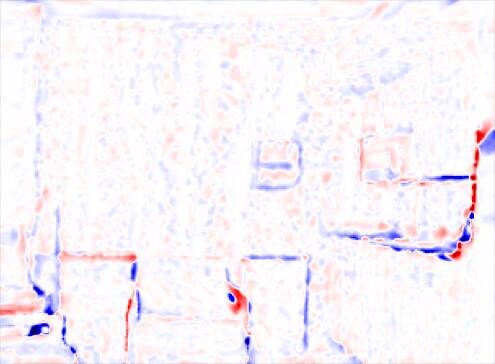}} & 
\framebox{\includegraphics[width=\figSize]{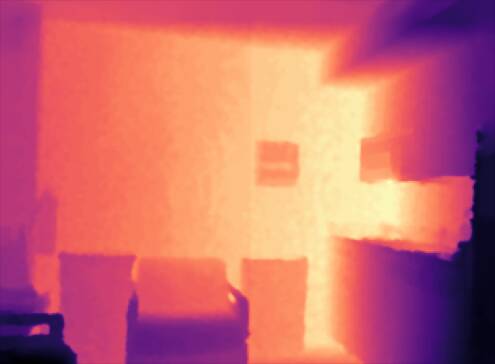}} \\
\framebox{\includegraphics[width=\figSize]{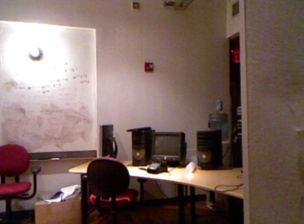}} &
\framebox{\includegraphics[width=\figSize]{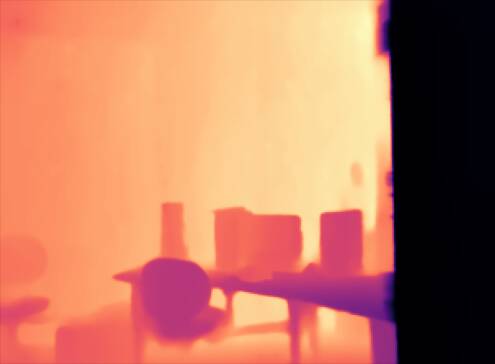}} &
\framebox{\includegraphics[width=\figSize]{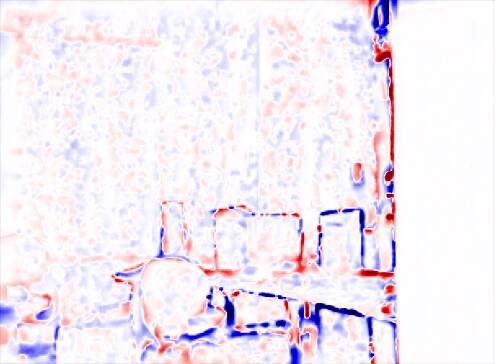}} &
\framebox{\includegraphics[width=\figSize]{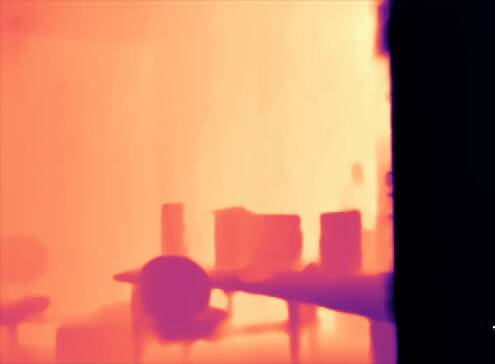}} &
\framebox{\includegraphics[width=\figSize]{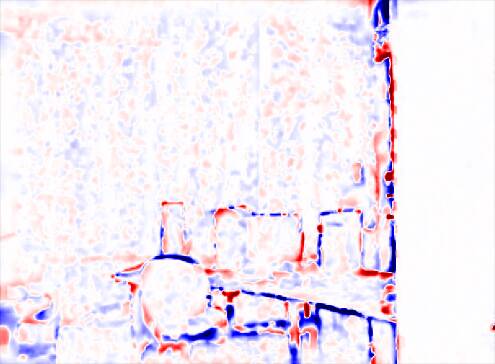}} &
\framebox{\includegraphics[width=\figSize]{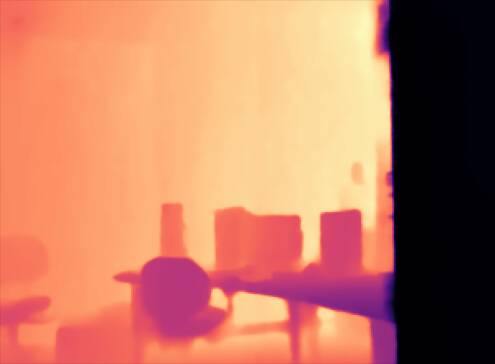}} &
\framebox{\includegraphics[width=\figSize]{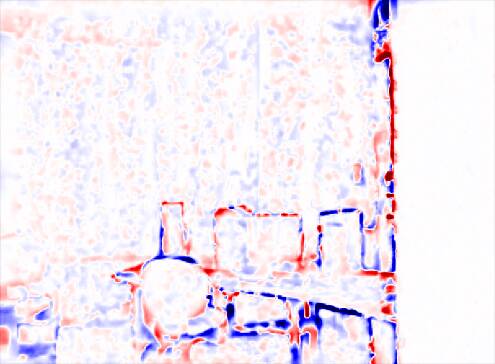}} & 
\framebox{\includegraphics[width=\figSize]{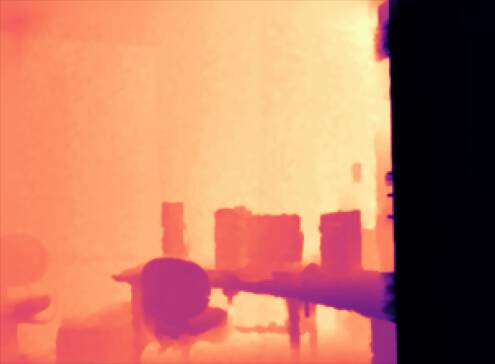}} \\
\framebox{\includegraphics[width=\figSize]{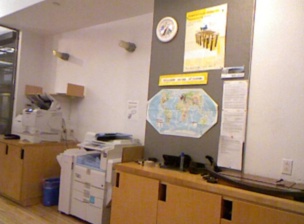}} &
\framebox{\includegraphics[width=\figSize]{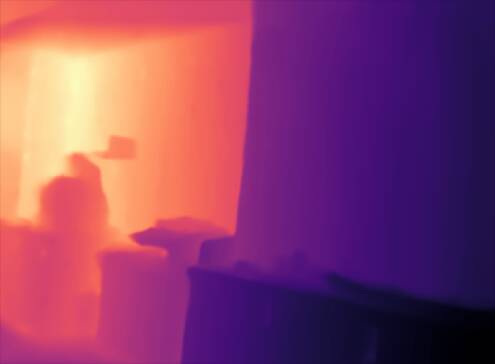}} &
\framebox{\includegraphics[width=\figSize]{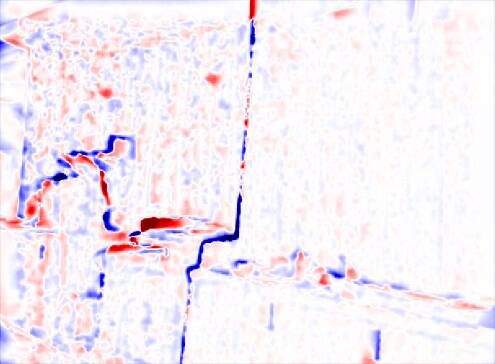}} &
\framebox{\includegraphics[width=\figSize]{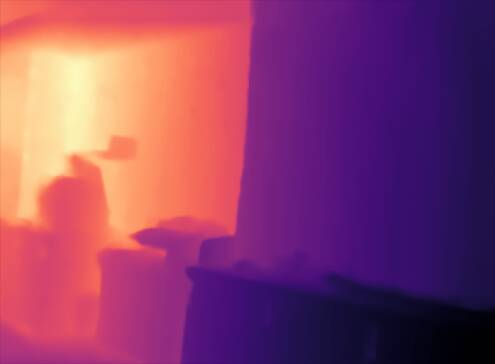}} &
\framebox{\includegraphics[width=\figSize]{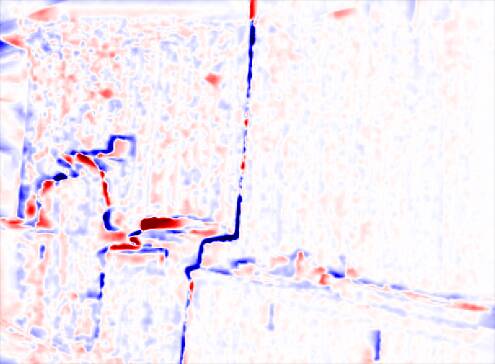}} &
\framebox{\includegraphics[width=\figSize]{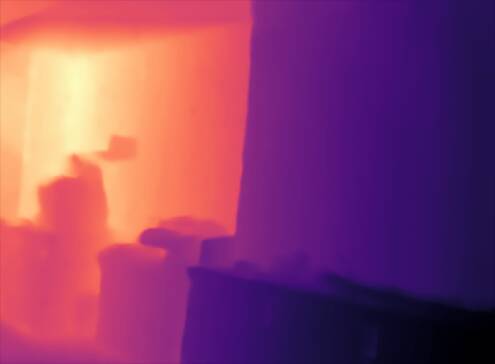}} &
\framebox{\includegraphics[width=\figSize]{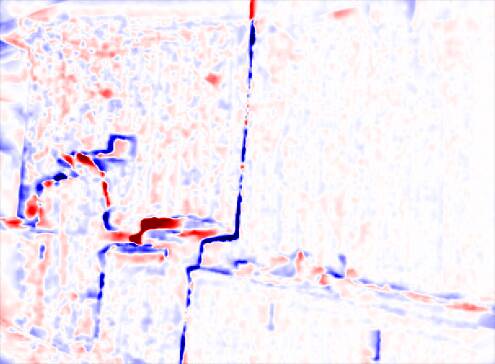}} & 
\framebox{\includegraphics[width=\figSize]{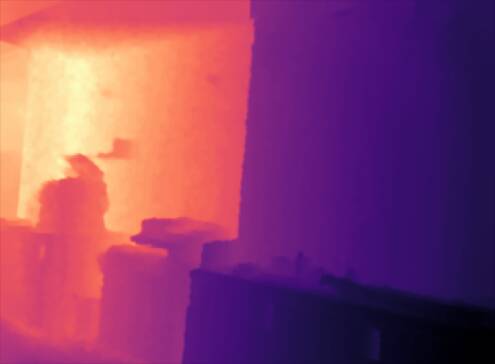}} \\
\specialrule{\arrayrulewidth}{2mm}{0mm}
\rowcolor{blue!10}\multicolumn{8}{c}{\bf \sdsdata~($640 \times 480$)}\\
\specialrule{\arrayrulewidth}{0mm}{2mm}
\framebox{\includegraphics[width=\figSize]{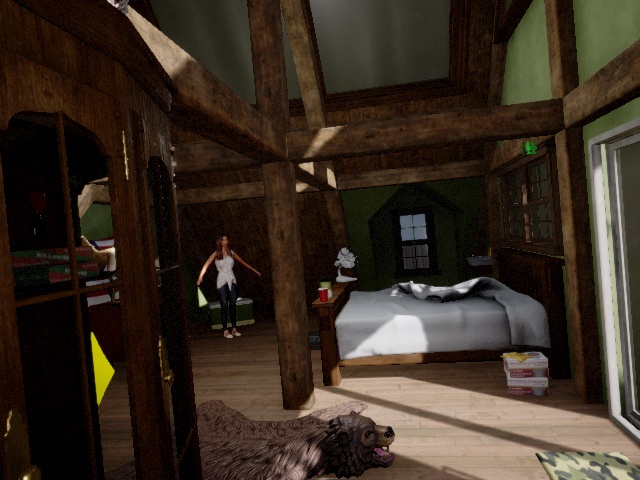}} &
\framebox{\includegraphics[width=\figSize]{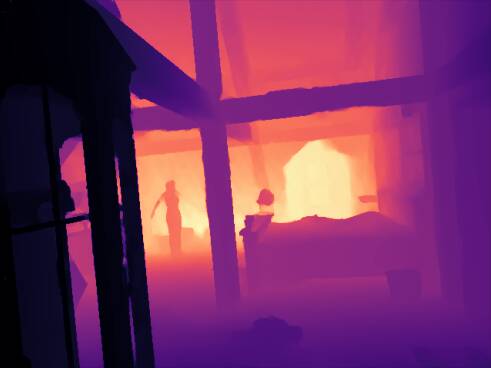}} &
\framebox{\includegraphics[width=\figSize]{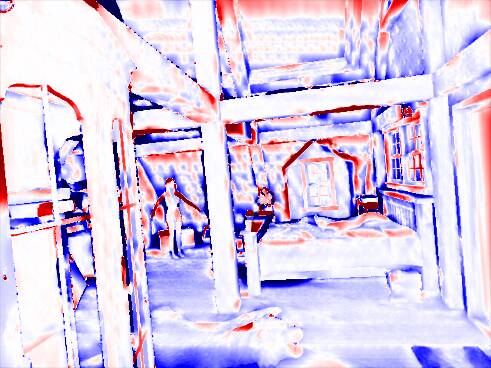}} &
\framebox{\includegraphics[width=\figSize]{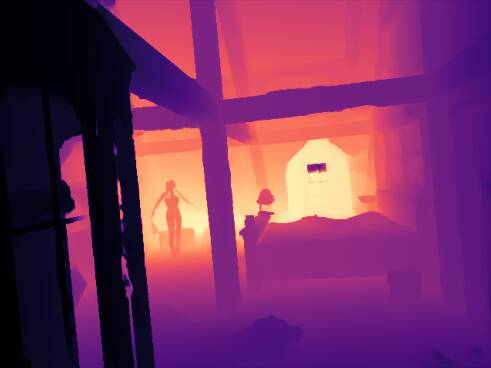}} &
\framebox{\includegraphics[width=\figSize]{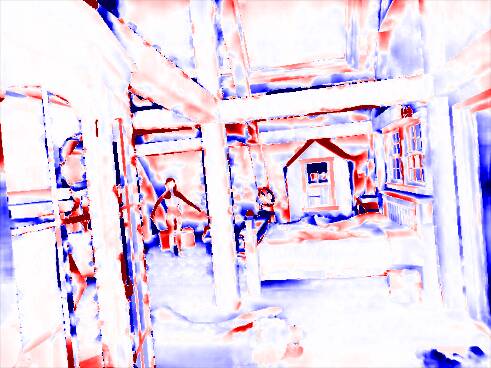}} &
\framebox{\includegraphics[width=\figSize]{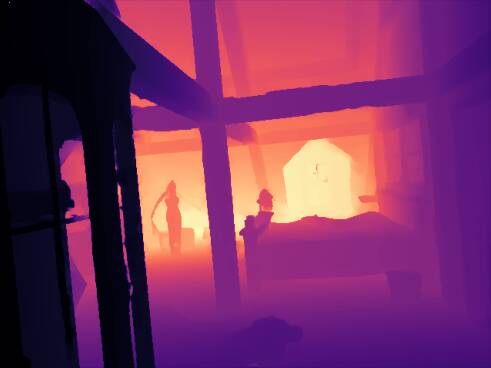}} &
\framebox{\includegraphics[width=\figSize]{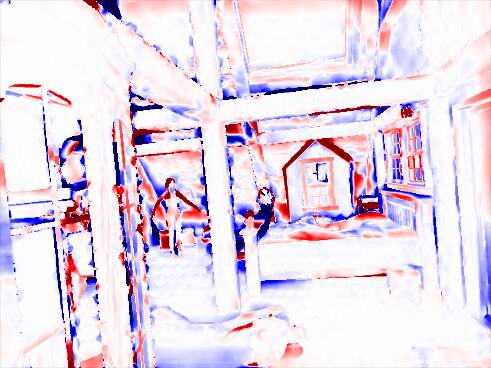}} & \framebox{\includegraphics[width=\figSize]{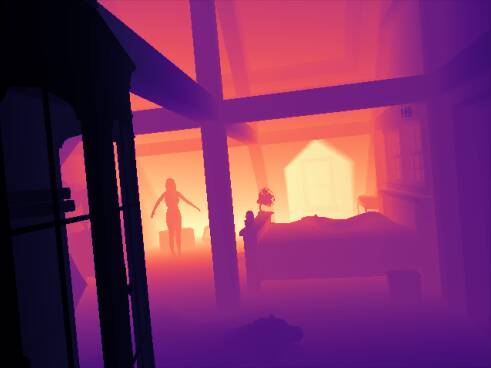}} \\
\framebox{\includegraphics[width=\figSize]{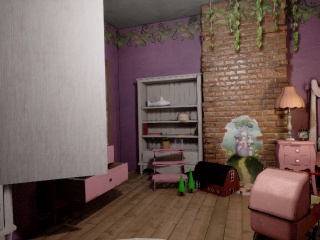}} &
\framebox{\includegraphics[width=\figSize]{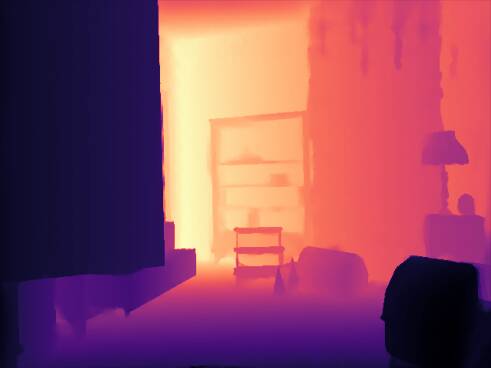}} &
\framebox{\includegraphics[width=\figSize]{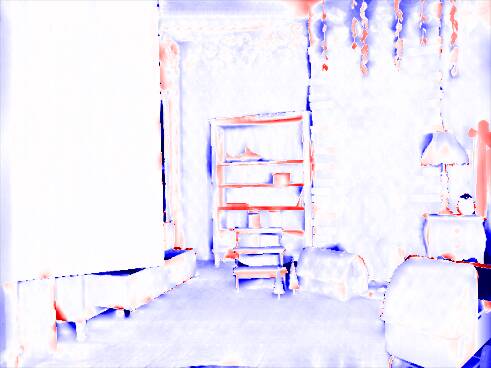}} &
\framebox{\includegraphics[width=\figSize]{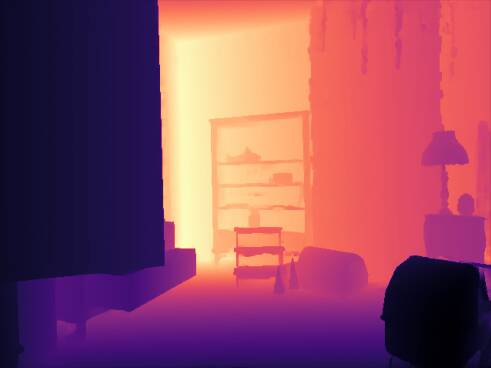}} &
\framebox{\includegraphics[width=\figSize]{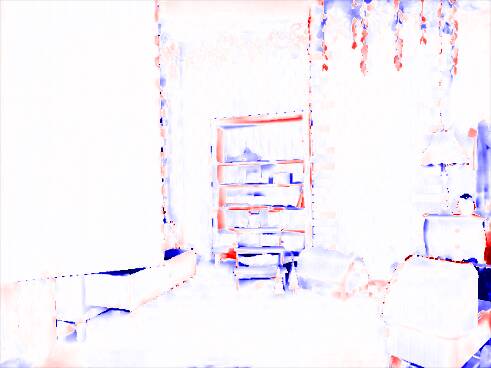}} &
\framebox{\includegraphics[width=\figSize]{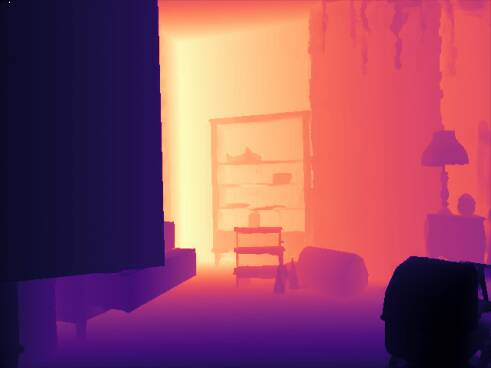}} &
\framebox{\includegraphics[width=\figSize]{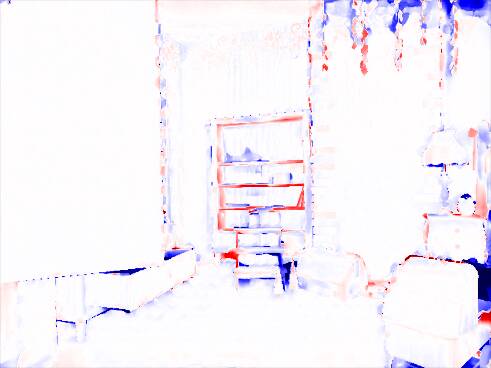}} & \framebox{\includegraphics[width=\figSize]{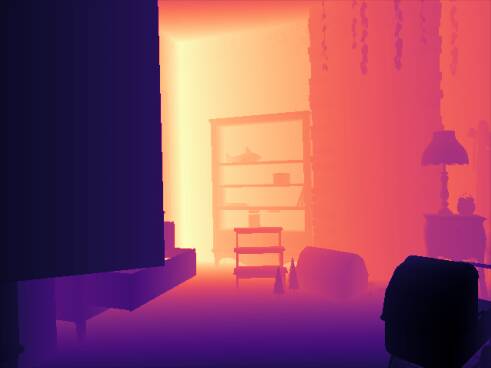}} \\
\framebox{\includegraphics[width=\figSize]{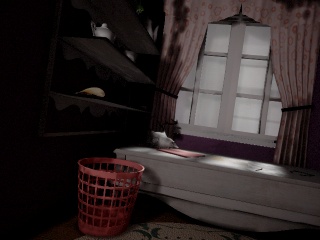}} &
\framebox{\includegraphics[width=\figSize]{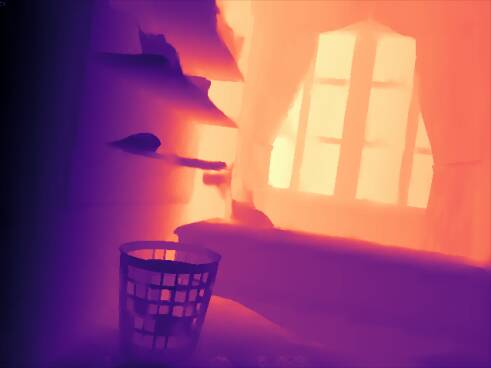}} &
\framebox{\includegraphics[width=\figSize]{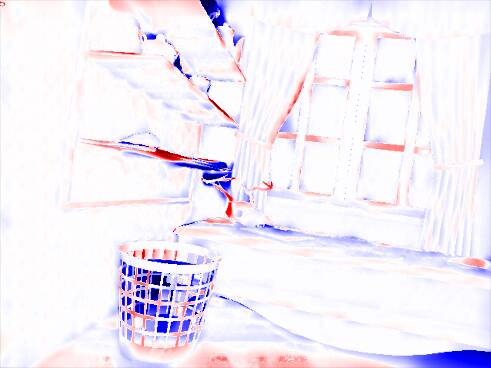}} &
\framebox{\includegraphics[width=\figSize]{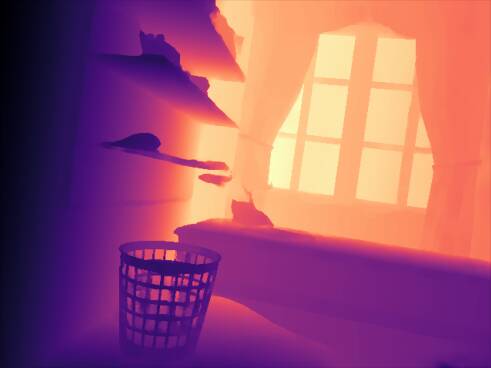}} &
\framebox{\includegraphics[width=\figSize]{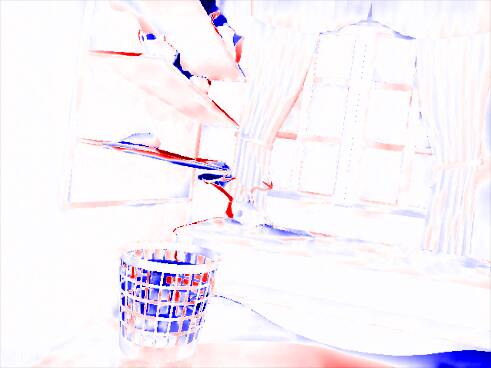}} &
\framebox{\includegraphics[width=\figSize]{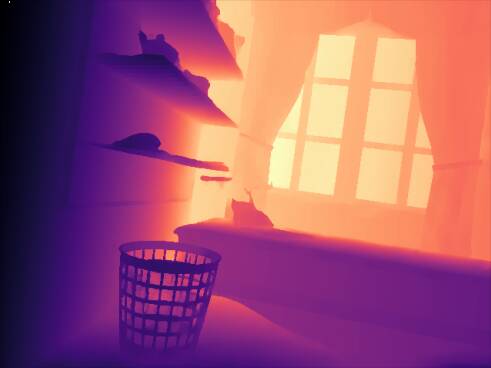}} &
\framebox{\includegraphics[width=\figSize]{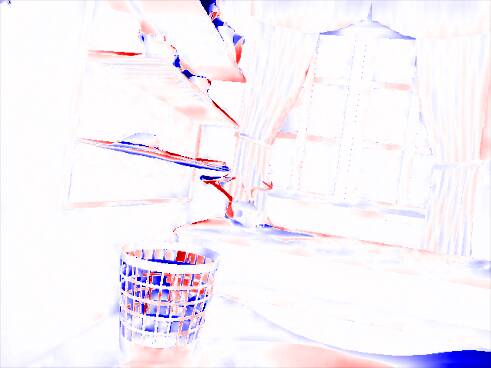}} & \framebox{\includegraphics[width=\figSize]{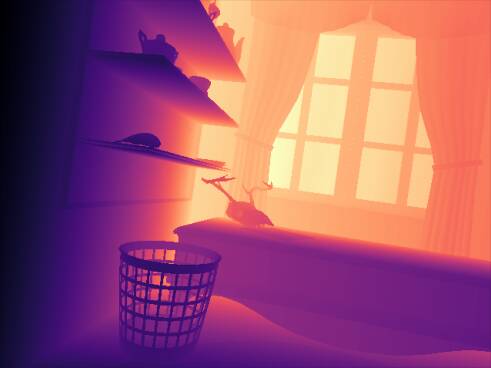}} 
\end{tabular}
\caption{Qualitative results. We report, for three arbitrary frames in the test sets of NYU-Depth v2 (top rows) and \sdsdata~(bottom rows), the predicted depth maps $\hat{D}$ and error maps $\hat{D}-D_{\rm GT}$ (range: $[-500, 500] mm$). Figure best viewed in color.}
\label{fig:results-q}
\vspace{-3mm}
\end{figure*}
\label{sec:perf_noquantization}

We first discuss some quantitative results of our method (in $\rm float32$ precision) against others in \cref{tab:results-dc}. We achieve very good results on both evaluated datasets despite their different sparsity level $K$: our method ranks \underline{second-best} on NYU-Depth v2 with a RMSE of $63.7 \, mm$ and {\bf best} on \sdsdata{} with a RMSE of $98.7 \, mm$. MAE and MNS confirm the RMSE ranking.  We also achieve very competitive $t_{\rm GPU}$ by virtue of the simple initialization obtained via NNI and simple encoder-decoder architecture of \cref{fig:networkandtraining}. 

NLSPN provides the most competitive performances: it is quality-wise the {\bf best} approach on NYU-Depth v2 (but the RMSE gap w.r.t. ours is very limited, only $2.3 \, mm$) and \underline{second-best} on \sdsdata, as confirmed by all quality metrics as well as visual results (\cref{sec:visual}). However, its $t_{\rm GPU}$ is relatively high even if the model is substantially compact. This is due to spatial propagation, which requires several iterations (18 as recommended in \cite{park_non-local_2020}) that linearly increase $t_{\rm GPU}$ even at constant memory footprint. $t_{\rm GPU}$ also notably increases significantly between $304 \times 224$ and $640 \times 480$ resolutions. Moreover, the initialization cost (\ie, the input encoder-decoder module of NLSPN) is substantially higher than that of our method, which runs on CPU. 

As for the other methods, it is possible to see that CSPN has slightly but consistently inferior performances than NLSPN and ours on both datasets. We also observed that S2D, even if very fast, yields sub-optimal quality metrics on NYU-Depth v2, and poor quality metrics on \sdsdata{} given its larger resolution at lower $K$. This confirms the findings in \cite[Sec. 6.4]{ma_self-supervised_2019} that the RMSE of S2D deteriorates for lower sparsity level $K$ and that structured subsampling patterns yield worse performances than random ones. Finally, we observed that $\rm D^3$, even if starting from the same initialization as ours, yields worse RMSE and MAE.

\subsection{Quantized Network Models}
\label{sec:perf_quantization}
The QAT results for our MP models are given in~\cref{tab:results-mp}. Among those models, we highlight our best weights-only MP QAT result for $\rm W_\star$; the weights size constraint of $14 \, \rm MB$ is met by weights memory size of $13.9 \, \rm MB$ after training. This yields $14$-fold memory size reduction w.r.t. the $\rm float32$ model at $198.5 \, \rm MB$. There, we observe limited and graceful degradation of the quality metrics\footnote{We do not report $t_{\rm GPU}$ as comparable with that of the $\rm float32$ model and not indicative of computation time on dedicated hardware, since fixed-precision operations are here emulated by $\rm float32$ operations as in \cite{nikolic2020bitpruning, uhlich_mixed_2020}.} (\eg, the RMSE loss is only $1.2$ mm on NYU-Depth v2) when reducing precision of the weights and activations in the $\rm float32$ model. 
However, as activations are also crucial for reducing memory footprint (\cref{sec:activations}), we remark that the overall most compact model with minimal impact on quality metrics is $\rm W_4 \rm A_8$ with a RMSE loss of only $0.7$ mm on NYU-Depth v2, as $\rm W_4 A_4$ sees instead significant degradation of the MRE, MNS, and MAE metrics (\eg, $6.6$ mm in RMSE). We select this model for the following analysis.

\subsection{Qualitative Analysis}
\label{sec:visual}
We now highlight the visual differences observed between different depth completion approaches. We consider the best competitor from \cref{sec:arch}, NLSPN, against our method in its ${\rm float32}$ and MP $\rm W_4 A_8$ flavors. We report three samples from each dataset with both depth maps $\hat{D}$ and error maps $\hat{D}-D_{\rm GT}$ in the same range to allow for a visual comparison. On NYU-Depth v2 NLSPN yields sharp edges with minimal amount of ``mixed depth'' pixels. However, the $\text{float32}$ and $\rm W_4 A_8$ models are as sharp, \eg, on the wall discontinuities (middle row). The error patterns seen in all approaches on far planar regions (top row) are probably due to the acquisition of $D_{\rm GT}$ with a Kinect sensor, whose accuracy decreases with the distance~\cite{zanuttigh_time--flight_2016}.

As for the higher-resolution \sdsdata~dataset, we see that while depth map quality of our models is comparable to that of NLSPN, on several scenes the latter yields larger errors on planar surfaces (\eg, the ground plane) than ours as shown in the corresponding error maps (top, middle row); our model also recovers more accurately some complex object details such as the cupboard (middle row), trash bin and shelves (bottom row).

\section{Conclusion}
\label{sec:conclusion}
We described a depth completion model based on a compact CNN with mixed-precision quantization. Our model performs within a few percent units of the state-of-the-art in terms of quality metrics on standard datasets such as NYU-Depth v2, but achieves faster GPU time at competitively small memory footprint and is suitable for real-time applications. A dedicated hardware implementation of our QNN will take full advantage of mixed-precision by fixed-point operations. We will investigate knowledge distillation with MP QAT to compress the network, as well as domain adaptation techniques \cite{agresti_2021, lopez-rodriguez_project_2020} to improve depth map quality on real RGB-ToF datasets. 
{
    \small
    \bibliographystyle{ieee_fullname}
    \bibliography{main}
}
\end{document}


\title{A Low-Footprint Quantized Neural Network for \\ Depth Completion of Very Sparse Time-of-Flight Depth Maps\\
\textit{Supplementary Material}}
\author{
    Xiaowen Jiang\textsuperscript{1}
    \quad
    Valerio Cambareri\textsuperscript{2}\thanks{Corresponding author:~\url{valerio.cambareri@sony.com}.}
    \quad
    Gianluca Agresti\textsuperscript{3}
    \\
    Cynthia Ifeyinwa Ugwu\textsuperscript{4}
    \quad
    Adriano Simonetto\textsuperscript{4}
    \quad
    Fabien Cardinaux\textsuperscript{3}
    \quad
    Pietro Zanuttigh\textsuperscript{4}\vspace{0.5ex} \\ 
	\textsuperscript{1} EPFL, Switzerland
    \quad \textsuperscript{2} Sony Depthsensing Solutions NV, Belgium\\
    \quad \textsuperscript{3} Sony Europe B.V., R\&D Center, Stuttgart Laboratory 1, Germany
    \quad \textsuperscript{4} University of Padova, Italy
}
\maketitle
\thispagestyle{empty}
\section{Sparse ToF Datasets}
\begin{figure*}[t]
    \centering
    \setlength{\fboxsep}{0pt}
    \subfloat[$1216 \times 352, K \approx 4.4 \%$]{
        \framebox{\includegraphics[width=0.99\textwidth]{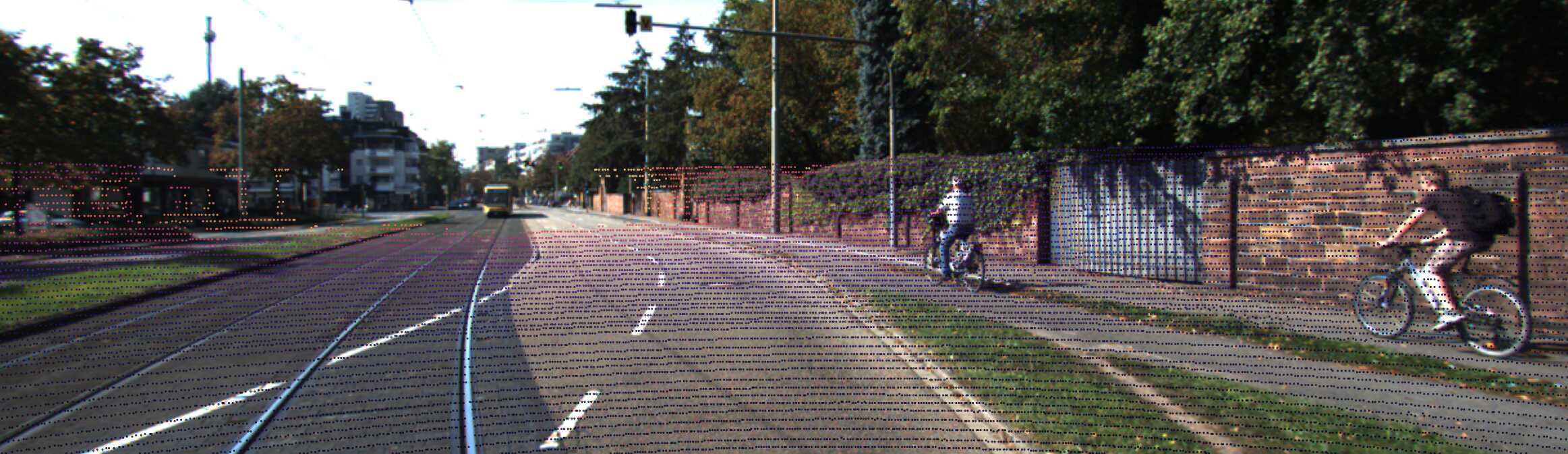}}
    }\\
    \vspace{1ex}
    \null\hfill
    \subfloat[\label{fig:nyuex}$304 \times 224, K \approx 1.4 \%$]{
        \framebox{\includegraphics[height=0.35\textwidth]{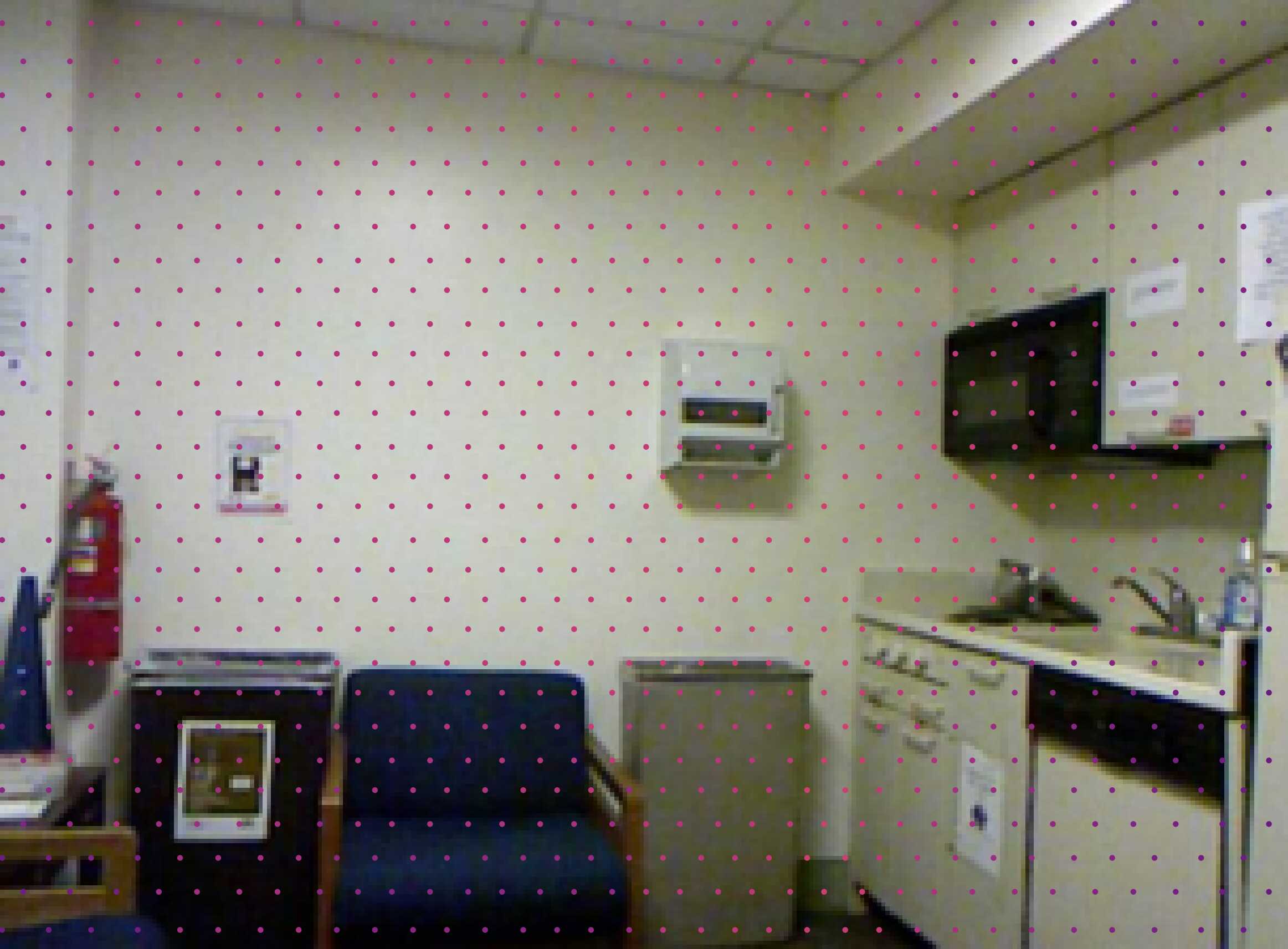}}
    }
    \hspace{3ex}
    \subfloat[\label{fig:sdsex}$640 \times 480, K \approx 0.4 \%$]{
        \framebox{\includegraphics[height=0.35\textwidth]{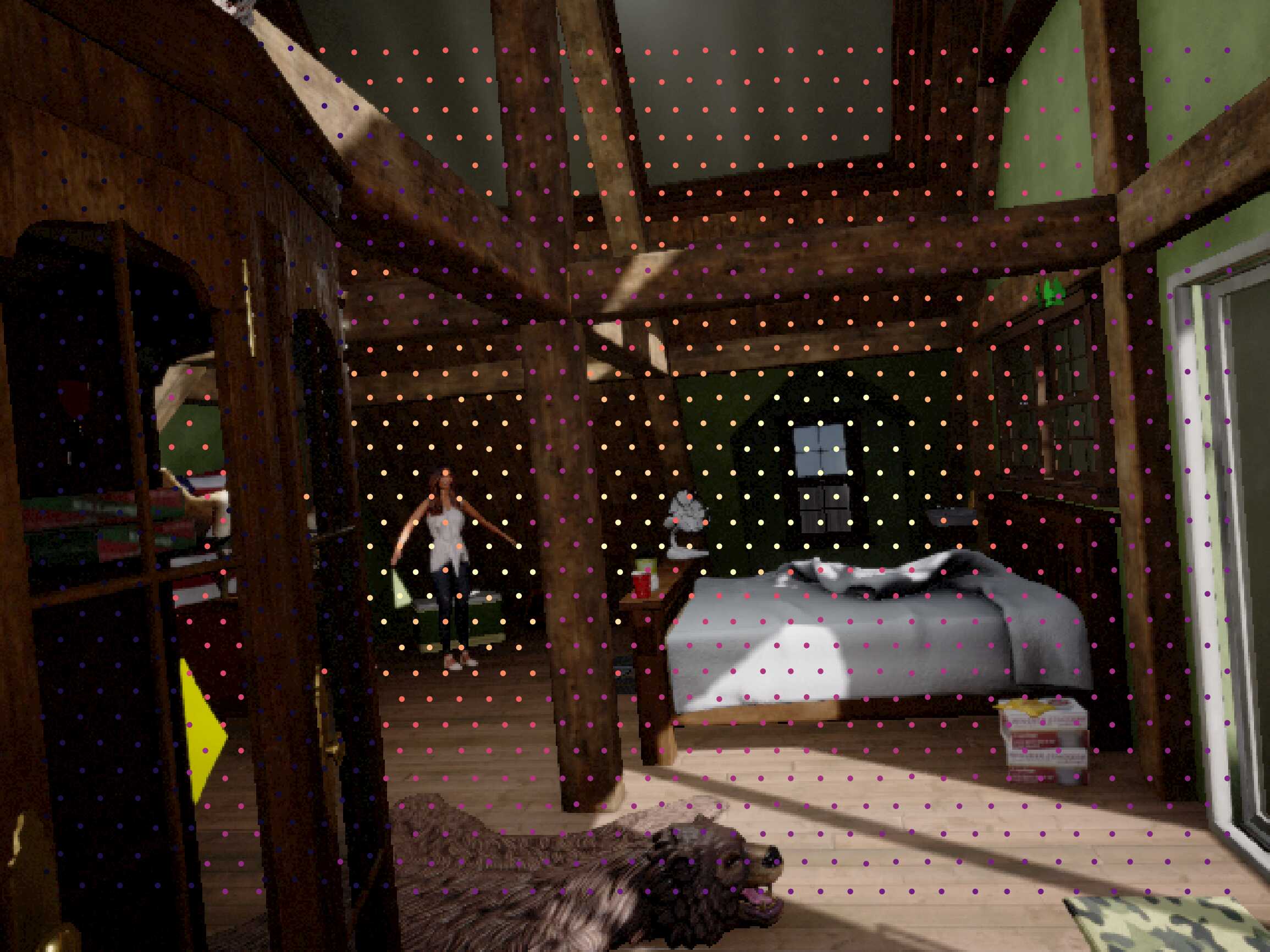}}
    }
    \hfill\null
    \caption{RGB-D overlay of (a) KITTI (LiDAR; range: $[0, 85] m$), (b) NYU-Depth v2 (processed; range: $[0, 10] m$), (c) \sdsdata~(sparse ToF; range: $[0, 15] m$). Depth overlay in ``magma'' colormap. Figure best viewed in color.
    \label{fig:kittivsdataset}}
\end{figure*}

We report an enlarged version of the pictures in the manuscript detailing our sparse ToF datasets in \cref{fig:kittivsdataset}. In \sdsdata, we simulate the projection of the dot pattern via a raytraced light shading profile defined parametrically and emulating a commercial VCSEL illuminator. We then receive the simulated sensor plane irradiance, and generate the ToF sensor pixel response. The rays corresponding to dot center locations on the sensor are annotated, and the sparse depth map retrieved  accordingly. We can observe parallax effects on the dot pattern due to optically accurate simulation, \eg, on the wooden beams to the left of the field of view in \cref{fig:sdsex}. As for NYU-Depth v2, we process the depth maps by \emph{masking} with a dot pattern that does not account for scene depth (\ie, it is not projected on the scene, but generated by assuming a default arbitrarily far plane). A sample from the KITTI dataset with Velodyne LiDAR overlay is also reported to compare visually the dot pattern density; there we can also observe that the sparse depth samples are not equally distributed over the RGB frame (the top part of which typically does not yield meaningful predictions as it includes, \eg, the sky).

\section{Loss Function}
\begin{figure*}[t]
    \centering
    \subfloat[Scale-dependent loss term]{
        \begin{tikzpicture}
\begin{axis}[
width=5in,
height=2.5in,
cycle list/Dark2,
xmin=0,
xmax=39,
xlabel={Epoch},
xlabel style={
font=\scriptsize},
ymode=log,
ymin=1e-3,
ylabel={$\mathcal{L}_{\ell_1}$},
ylabel style={
at = {(-0.08,0.5)},
font=\scriptsize},
axis background/.style={fill=white},
legend style={
legend columns=2, 
legend cell align=left,
align=left,
fill=white, fill opacity=0.65,
text opacity=1,
draw=none,
font=\scriptsize\selectfont
},
legend image code/.code={
\draw[mark repeat=2,mark phase=2]
plot coordinates {
(0cm,0cm)
(0.1cm,0cm)        
(0.2cm,0cm)         
};%
}
]
\addplot +[thick] table [x=Step, y=Value, col sep=comma] {supplementary/data/run-log_20210803-143647_configSpotNet_NNedt_L1_cossim_40_8_validation-tag-epoch_depth_loss.csv};
\addlegendentry{$\lambda_{\rm _n} = 1$}
\addplot +[thick] table [x=Step, y=Value, col sep=comma] {supplementary/data/run-log_20210730-165306_configSpotNet_NNedt_L1_1e-1xcossim_40_8_validation-tag-epoch_depth_loss.csv};
\addlegendentry{$\lambda_{\rm _n} = 10^{-1}$}
\addplot +[thick] table [x=Step, y=Value, col sep=comma] {supplementary/data/run-log_20210730-164931_configSpotNet_NNedt_L1_1e-2xcossim_40_8_validation-tag-epoch_depth_loss.csv};
\addlegendentry{$\lambda_{\rm _n} = 10^{-2}$}
\addplot +[thick] table [x=Step, y=Value, col sep=comma] {supplementary/data/run-log_20210727-155320_configBestSpotNet_NNedt_L1_1e-3xcossim_40_8_validation-tag-epoch_depth_loss.csv};
\addlegendentry{$\lambda_{\rm _n} = 10^{-3}$}
\addplot +[thick] table [x=Step, y=Value, col sep=comma] {supplementary/data/run-log_20210730-164656_configSpotNet_NNedt_L1_1e-4xcossim_40_8_validation-tag-epoch_depth_loss.csv};
\addlegendentry{$\lambda_{\rm _n} = 10^{-4}$}

\addplot +[mark=square, color=red] coordinates {(32,  0.001607)};
\addlegendentry{Best Model}
\end{axis}
\end{tikzpicture}
    }\\
    \subfloat[Scale-independent loss term]{
        \begin{tikzpicture}
\begin{axis}[
width=5in,
height=2.5in,
cycle list/Dark2,
xmin=0,
xmax=39,
xlabel={Epoch},
xlabel style={
font=\scriptsize},
ymin=-0.9,
ymax=-0.5,
ylabel={$\mathcal{L}_{\rm n}$},
ylabel style={
at = {(-0.08,0.5)},
font=\scriptsize},
axis background/.style={fill=white},
legend style={
legend columns=2, 
legend cell align=left,
align=left,
fill=white, fill opacity=0.65,
text opacity=1,
draw=none,
font=\scriptsize\selectfont
},
legend image code/.code={
\draw[mark repeat=2,mark phase=2]
plot coordinates {
(0cm,0cm)
(0.1cm,0cm)        
(0.2cm,0cm)         
};%
}
]
\addplot +[thick] table [x=Step, y=Value, col sep=comma] {supplementary/data/run-log_20210803-143647_configSpotNet_NNedt_L1_cossim_40_8_validation-tag-epoch_normals_loss.csv};
\addlegendentry{$\lambda_{\rm _n} = 1$}
\addplot +[thick] table [x=Step, y=Value, col sep=comma] {supplementary/data/run-log_20210730-165306_configSpotNet_NNedt_L1_1e-1xcossim_40_8_validation-tag-epoch_normals_loss.csv};
\addlegendentry{$\lambda_{\rm _n} = 10^{-1}$}
\addplot +[thick] table [x=Step, y=Value, col sep=comma] {supplementary/data/run-log_20210730-164931_configSpotNet_NNedt_L1_1e-2xcossim_40_8_validation-tag-epoch_normals_loss.csv};
\addlegendentry{$\lambda_{\rm _n} = 10^{-2}$}
\addplot +[thick] table [x=Step, y=Value, col sep=comma] {supplementary/data/run-log_20210727-155320_configBestSpotNet_NNedt_L1_1e-3xcossim_40_8_validation-tag-epoch_normals_loss.csv};
\addlegendentry{$\lambda_{\rm _n} = 10^{-3}$}
\addplot +[thick] table [x=Step, y=Value, col sep=comma] {supplementary/data/run-log_20210730-164656_configSpotNet_NNedt_L1_1e-4xcossim_40_8_validation-tag-epoch_normals_loss.csv};
\addlegendentry{$\lambda_{\rm _n} = 10^{-4}$}

\addplot +[mark=square, color=red] coordinates {(32, -0.76579)};
\addlegendentry{Best Model}
\end{axis}
\end{tikzpicture}
    }
    \caption{Loss Function Tuning on \sdsdata. We report the validation loss terms as $\lambda_{\rm n}$ varies: (a) scale-dependent term $\mathcal{L}_{\ell_1}$, (b) scale-independent term $\mathcal{L}_{\rm n}$. Figure best viewed in color.}
    \label{fig:lossdecay}
\end{figure*}
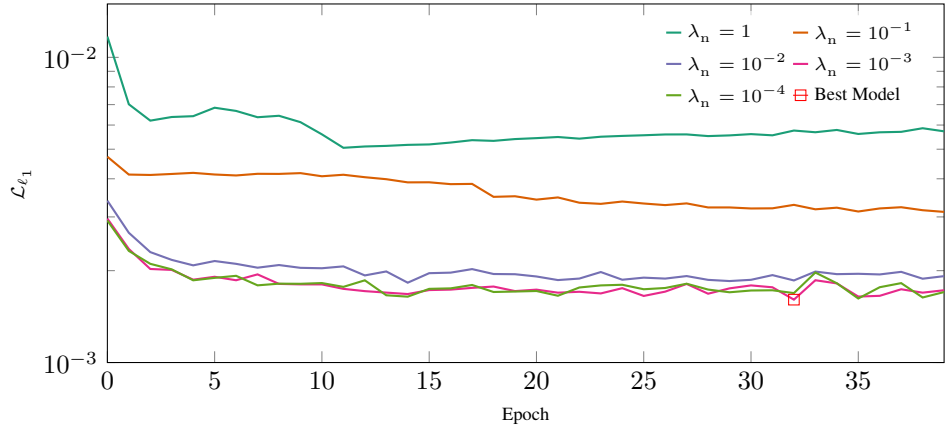
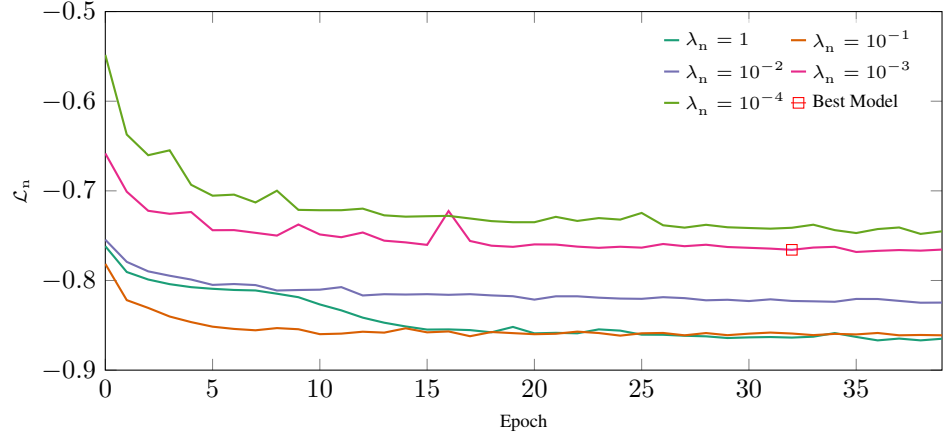

We here report an extended study of the choice of loss function parameter $\lambda_{\rm n}$ for the loss $\mathcal{L}(\mathcal{D})$ we utilize in the manuscript. The curves we report are validation loss curves of the scale-dependent term $\mathcal{L}_{\ell_1}$ and scale-independent term $\mathcal{L}_{\rm n}$ as the epochs increase. As we may observe in \cref{fig:lossdecay} high values of $\lambda_{\rm n}$ overly promote normals similarity at the cost of higher depth error in the $\ell_1$ sense; conversely, for lower values of the hyperparameter we attain low depth error at acceptable mean normals similarity values. We therefore find our optimum, $\lambda^{\rm opt}_{\rm n} = 10^{-3}$, to be the best trade-off between those shown in this analysis.

\section{Qualitative Evaluation}
\begin{figure*}[t]
\newcommand{\figSize}{17ex}
\tabcolsep=0.1cm
\centering
\scriptsize
\begin{tabular}{cccccccc}
\trule
\rowcolor{gray!10} & \multicolumn{2}{c}{\bf NLSPN} &  \multicolumn{2}{c}{ \bf Ours ($\text{float32}$) } &  \multicolumn{2}{c}{ \bf Ours ($\rm W_4 A_8$) } & \\
\rowcolor{gray!10} \multirow{-2}{*}{\bf Color}  & \bf Depth Map & \bf Error Map & \bf Depth Map & \bf Error Map & \bf Depth Map & \bf Error Map & \multirow{-2}{*}{$D_{\rm GT}$}\\
%
\specialrule{\arrayrulewidth}{2mm}{0mm}
\rowcolor{blue!10}\multicolumn{8}{c}{\bf \sdsdata~($640 \times 480$)}\\
\specialrule{\arrayrulewidth}{0mm}{2mm}
\includegraphics[width=\figSize]{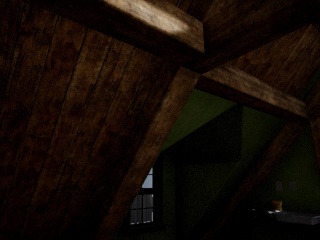} &
\includegraphics[width=\figSize,trim=2.08cm 1.28cm 1.6cm 1.4cm,clip]{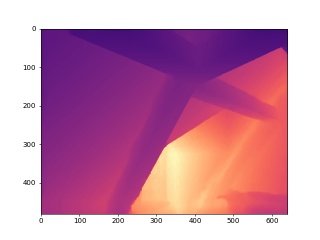} &
\includegraphics[width=\figSize,trim=2.08cm 1.28cm 1.6cm 1.4cm,clip]{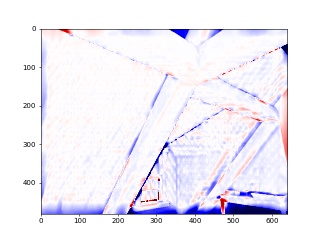} &
\includegraphics[width=\figSize,trim=2.08cm 1.28cm 1.6cm 1.4cm,clip]{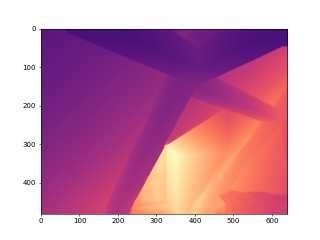} &
\includegraphics[width=\figSize,trim=2.08cm 1.28cm 1.6cm 1.4cm,clip]{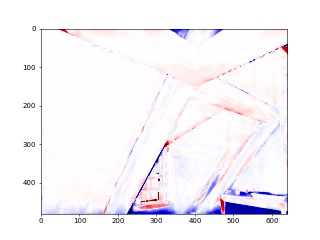} &
\includegraphics[width=\figSize,trim=2.08cm 1.28cm 1.6cm 1.4cm,clip]{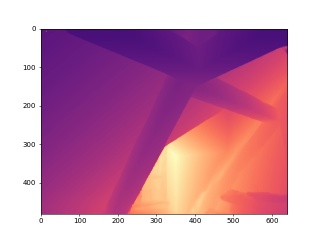} &
\includegraphics[width=\figSize,trim=2.08cm 1.28cm 1.6cm 1.4cm,clip]{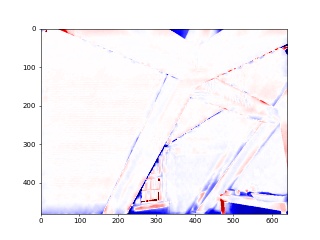} & \includegraphics[width=\figSize,trim=2.08cm 1.28cm 1.6cm 1.4cm,clip]{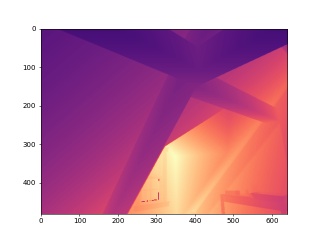} \\
\includegraphics[width=\figSize]{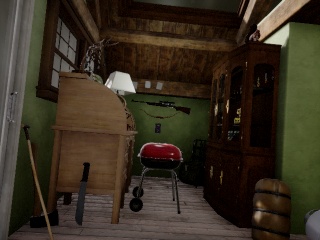} &
\includegraphics[width=\figSize,trim=2.08cm 1.28cm 1.6cm 1.4cm,clip]{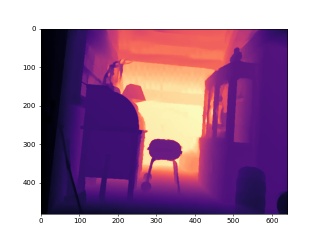} &
\includegraphics[width=\figSize,trim=2.08cm 1.28cm 1.6cm 1.4cm,clip]{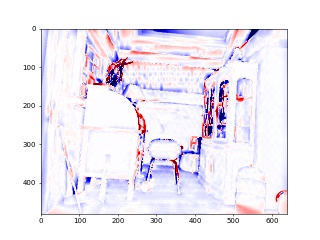} &
\includegraphics[width=\figSize,trim=2.08cm 1.28cm 1.6cm 1.4cm,clip]{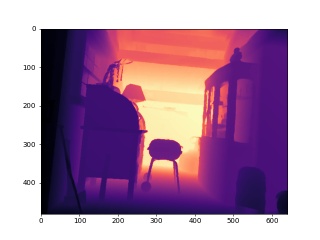} &
\includegraphics[width=\figSize,trim=2.08cm 1.28cm 1.6cm 1.4cm,clip]{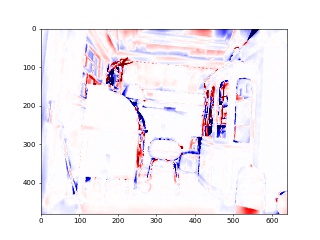} &
\includegraphics[width=\figSize,trim=2.08cm 1.28cm 1.6cm 1.4cm,clip]{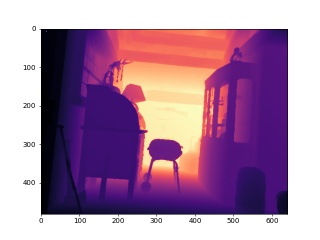} &
\includegraphics[width=\figSize,trim=2.08cm 1.28cm 1.6cm 1.4cm,clip]{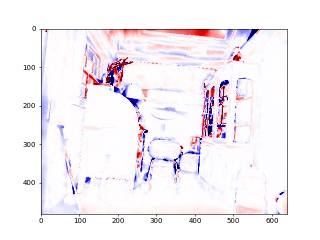} & \includegraphics[width=\figSize,trim=2.08cm 1.28cm 1.6cm 1.4cm,clip]{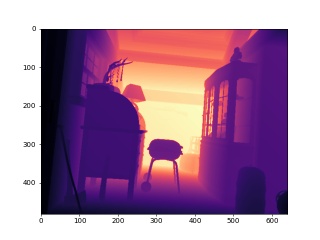} \\

\includegraphics[width=\figSize]{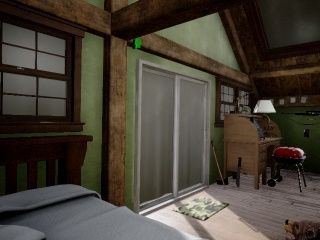} &
\includegraphics[width=\figSize,trim=2.08cm 1.28cm 1.6cm 1.4cm,clip]{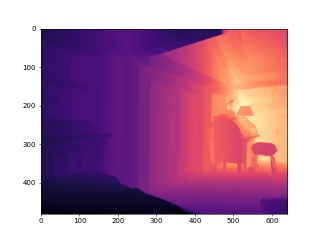} &
\includegraphics[width=\figSize,trim=2.08cm 1.28cm 1.6cm 1.4cm,clip]{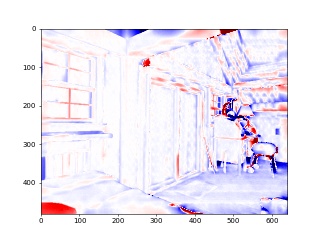} &
\includegraphics[width=\figSize,trim=2.08cm 1.28cm 1.6cm 1.4cm,clip]{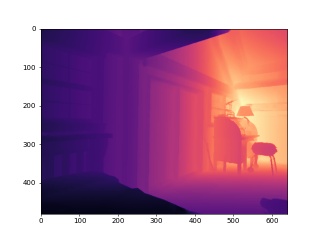} &
\includegraphics[width=\figSize,trim=2.08cm 1.28cm 1.6cm 1.4cm,clip]{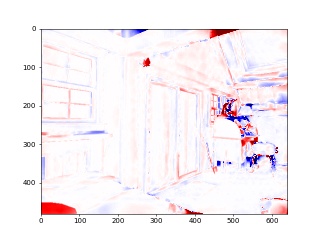} &
\includegraphics[width=\figSize,trim=2.08cm 1.28cm 1.6cm 1.4cm,clip]{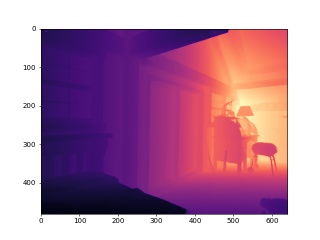} &
\includegraphics[width=\figSize,trim=2.08cm 1.28cm 1.6cm 1.4cm,clip]{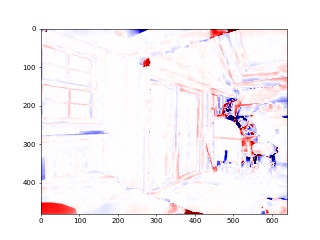} & 
\includegraphics[width=\figSize,trim=2.08cm 1.28cm 1.6cm 1.4cm,clip]{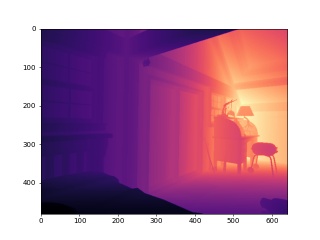} \\

\includegraphics[width=\figSize]{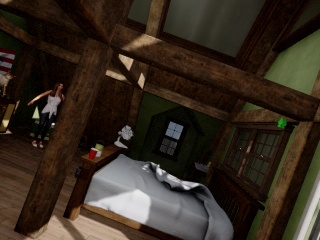} &
\includegraphics[width=\figSize,trim=2.08cm 1.28cm 1.6cm 1.4cm,clip]{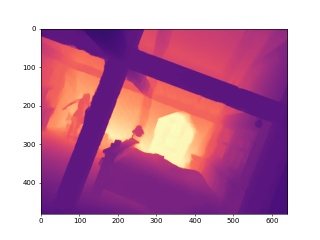} &
\includegraphics[width=\figSize,trim=2.08cm 1.28cm 1.6cm 1.4cm,clip]{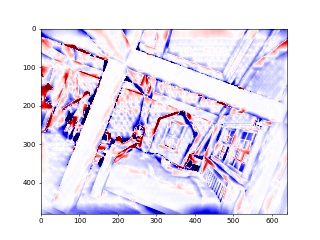} &
\includegraphics[width=\figSize,trim=2.08cm 1.28cm 1.6cm 1.4cm,clip]{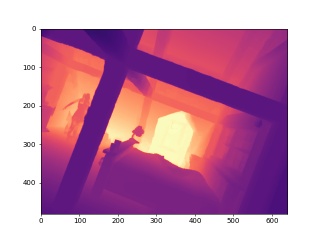} &
\includegraphics[width=\figSize,trim=2.08cm 1.28cm 1.6cm 1.4cm,clip]{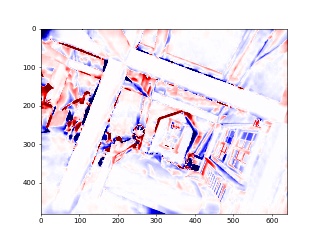} &
\includegraphics[width=\figSize,trim=2.08cm 1.28cm 1.6cm 1.4cm,clip]{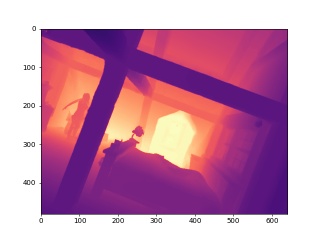} &
\includegraphics[width=\figSize,trim=2.08cm 1.28cm 1.6cm 1.4cm,clip]{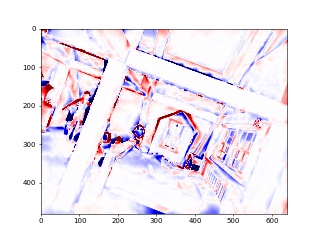} & 
\includegraphics[width=\figSize,trim=2.08cm 1.28cm 1.6cm 1.4cm,clip]{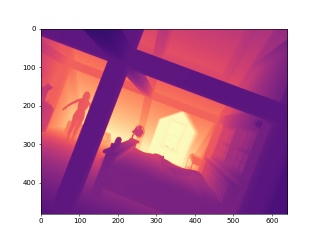} \\

\includegraphics[width=\figSize]{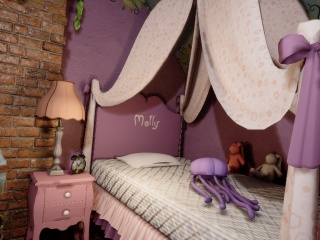} &
\includegraphics[width=\figSize,trim=2.08cm 1.28cm 1.6cm 1.4cm,clip]{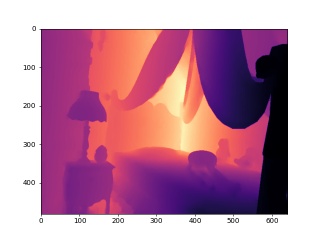} &
\includegraphics[width=\figSize,trim=2.08cm 1.28cm 1.6cm 1.4cm,clip]{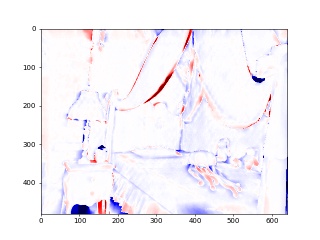} &
\includegraphics[width=\figSize,trim=2.08cm 1.28cm 1.6cm 1.4cm,clip]{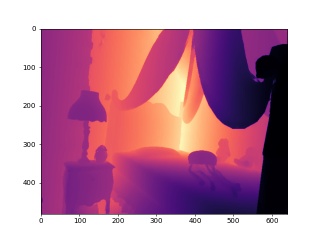} &
\includegraphics[width=\figSize,trim=2.08cm 1.28cm 1.6cm 1.4cm,clip]{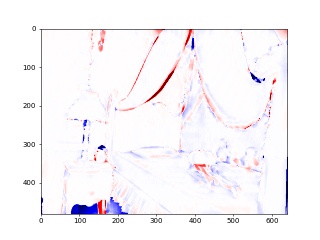} &
\includegraphics[width=\figSize,trim=2.08cm 1.28cm 1.6cm 1.4cm,clip]{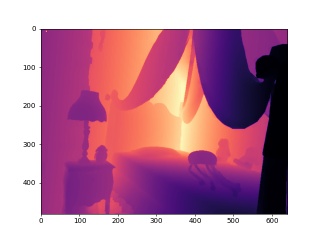} &
\includegraphics[width=\figSize,trim=2.08cm 1.28cm 1.6cm 1.4cm,clip]{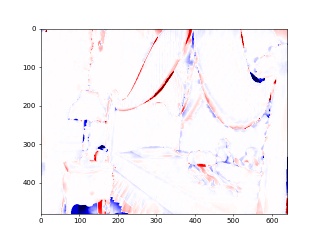} & \includegraphics[width=\figSize,trim=2.08cm 1.28cm 1.6cm 1.4cm,clip]{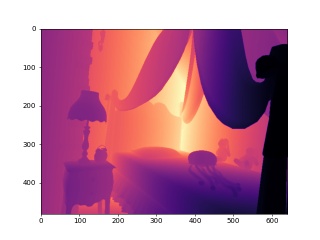}  \\

\includegraphics[width=\figSize]{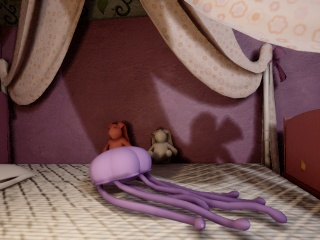} &
\includegraphics[width=\figSize,trim=2.08cm 1.28cm 1.6cm 1.4cm,clip]{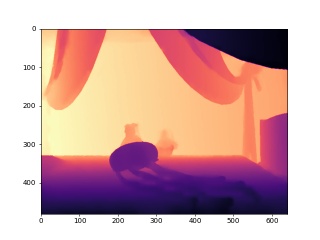} &
\includegraphics[width=\figSize,trim=2.08cm 1.28cm 1.6cm 1.4cm,clip]{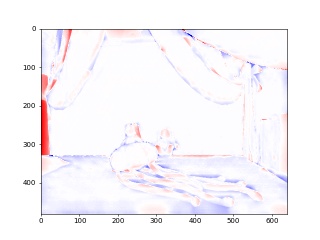} &
\includegraphics[width=\figSize,trim=2.08cm 1.28cm 1.6cm 1.4cm,clip]{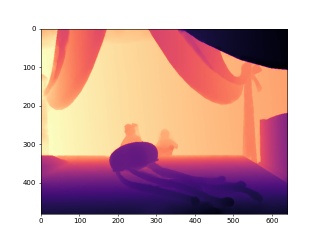} &
\includegraphics[width=\figSize,trim=2.08cm 1.28cm 1.6cm 1.4cm,clip]{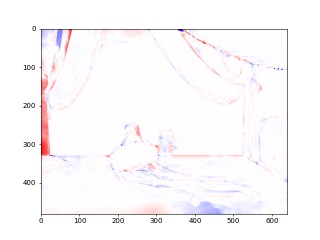} &
\includegraphics[width=\figSize,trim=2.08cm 1.28cm 1.6cm 1.4cm,clip]{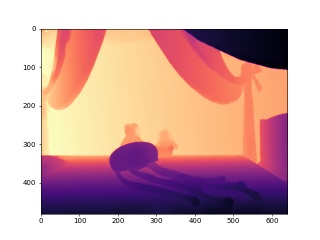} &
\includegraphics[width=\figSize,trim=2.08cm 1.28cm 1.6cm 1.4cm,clip]{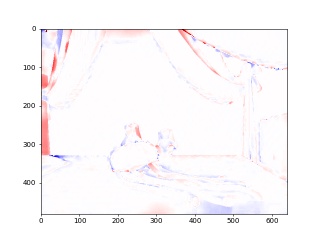} & \includegraphics[width=\figSize,trim=2.08cm 1.28cm 1.6cm 1.4cm,clip]{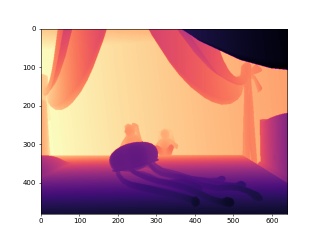} \\
\end{tabular}
\caption{Qualitative Results. We report, for several arbitrary frames in the test set of \sdsdata, the predicted depth maps $\hat{D}$ and error maps $\hat{D}-D_{\rm GT}$ (range: $[-500, 500] mm$). Figure best viewed in color.}
\label{fig:results_q_SDS_sup}
\end{figure*}

\begin{figure*}[t]
\newcommand{\figSize}{17ex}
\tabcolsep=0.1cm
\centering
\scriptsize
\begin{tabular}{cccccccc}
\trule
\rowcolor{gray!10} & \multicolumn{2}{c}{\bf NLSPN} &  \multicolumn{2}{c}{ \bf Ours ($\text{float32}$) } &  \multicolumn{2}{c}{ \bf Ours ($\rm W_4 A_8$) } & \\
\rowcolor{gray!10} \multirow{-2}{*}{\bf Color}  & \bf Depth Map & \bf Error Map & \bf Depth Map & \bf Error Map & \bf Depth Map & \bf Error Map & \multirow{-2}{*}{$D_{\rm GT}$}\\
\mruleu
\rowcolor{green!10}\multicolumn{8}{c}{\bf NYU-Depth v2 (center cropped, $304 \times 224$)}\\
\specialrule{\arrayrulewidth}{0mm}{2mm}
\includegraphics[width=\figSize]{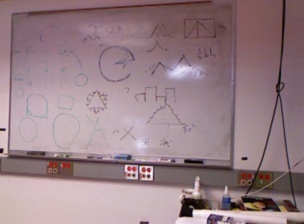} &
\includegraphics[width=\figSize,trim=2.03cm 1.34cm 1.6cm 1.5cm,clip]{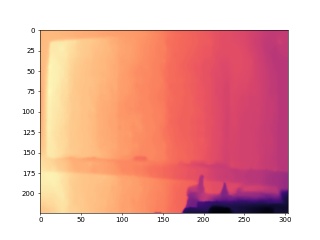} &
\includegraphics[width=\figSize,trim=2.03cm 1.34cm 1.6cm 1.5cm,clip]{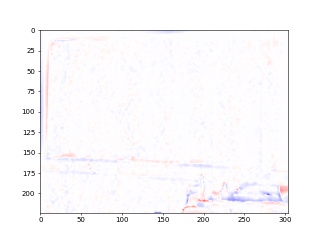} &
\includegraphics[width=\figSize,trim=2.03cm 1.34cm 1.6cm 1.5cm,clip]{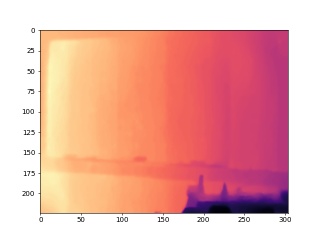} &
\includegraphics[width=\figSize,trim=2.03cm 1.34cm 1.6cm 1.5cm,clip]{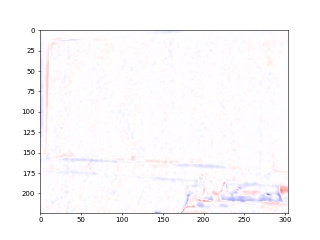} &
\includegraphics[width=\figSize,trim=2.03cm 1.34cm 1.6cm 1.5cm,clip]{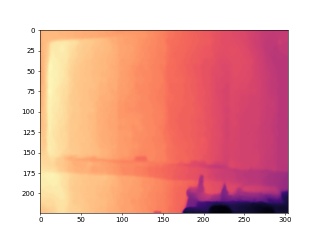} &
\includegraphics[width=\figSize,trim=2.03cm 1.34cm 1.6cm 1.5cm,clip]{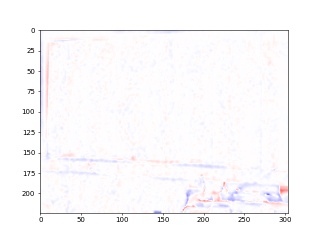} & 
\includegraphics[width=\figSize,trim=2.03cm 1.34cm 1.6cm 1.5cm,clip]{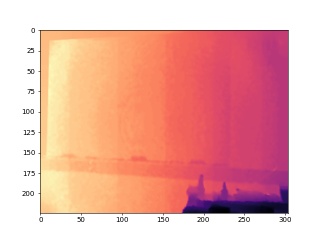} \\

\includegraphics[width=\figSize]{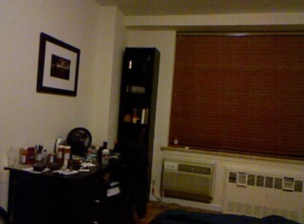} &
\includegraphics[width=\figSize,trim=2.03cm 1.34cm 1.6cm 1.5cm,clip]{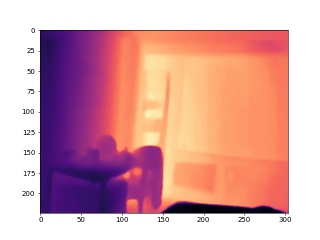} &
\includegraphics[width=\figSize,trim=2.03cm 1.34cm 1.6cm 1.5cm,clip]{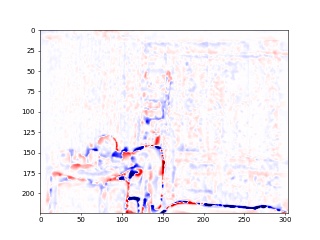} &
\includegraphics[width=\figSize,trim=2.03cm 1.34cm 1.6cm 1.5cm,clip]{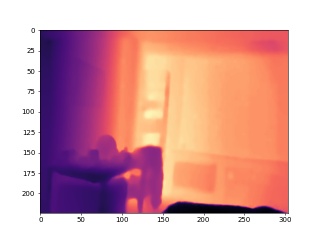} &
\includegraphics[width=\figSize,trim=2.03cm 1.34cm 1.6cm 1.5cm,clip]{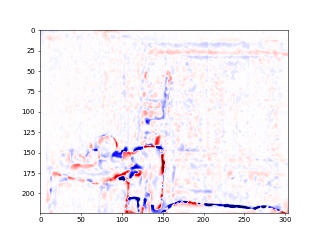} &
\includegraphics[width=\figSize,trim=2.03cm 1.34cm 1.6cm 1.5cm,clip]{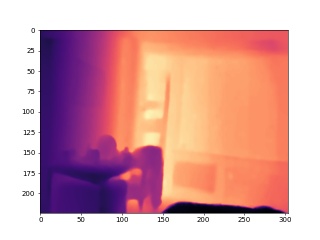} &
\includegraphics[width=\figSize,trim=2.03cm 1.34cm 1.6cm 1.5cm,clip]{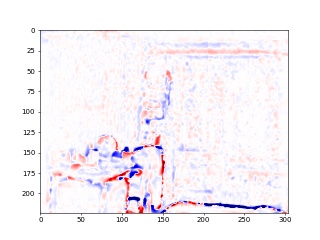} & 
\includegraphics[width=\figSize,trim=2.03cm 1.34cm 1.6cm 1.5cm,clip]{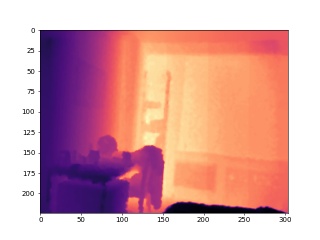} \\

\includegraphics[width=\figSize]{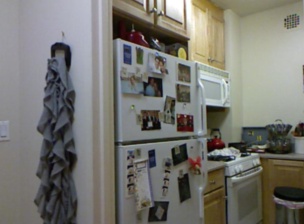} &
\includegraphics[width=\figSize,trim=2.03cm 1.34cm 1.6cm 1.5cm,clip]{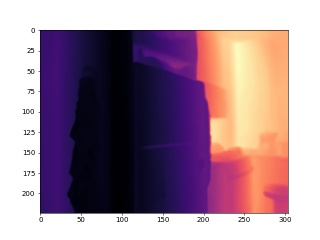} &
\includegraphics[width=\figSize,trim=2.03cm 1.34cm 1.6cm 1.5cm,clip]{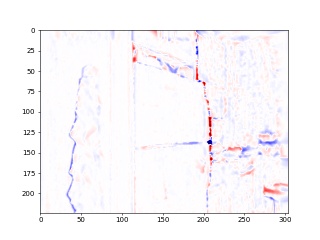} &
\includegraphics[width=\figSize,trim=2.03cm 1.34cm 1.6cm 1.5cm,clip]{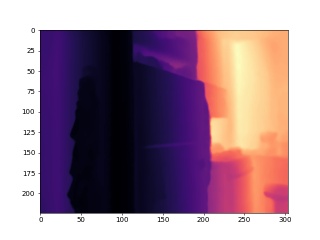} &
\includegraphics[width=\figSize,trim=2.03cm 1.34cm 1.6cm 1.5cm,clip]{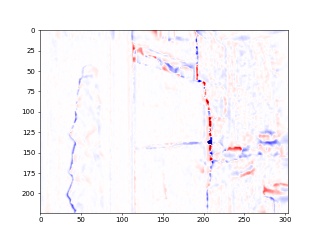} &
\includegraphics[width=\figSize,trim=2.03cm 1.34cm 1.6cm 1.5cm,clip]{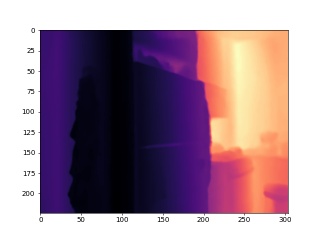} &
\includegraphics[width=\figSize,trim=2.03cm 1.34cm 1.6cm 1.5cm,clip]{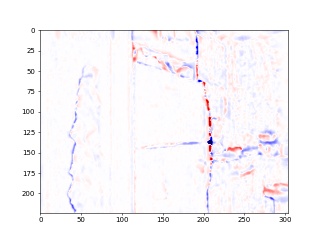} & 
\includegraphics[width=\figSize,trim=2.03cm 1.34cm 1.6cm 1.5cm,clip]{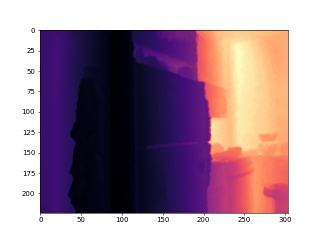} \\

\includegraphics[width=\figSize]{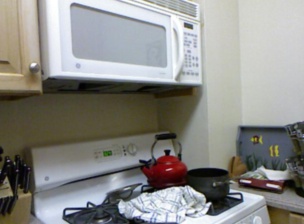} &
\includegraphics[width=\figSize,trim=2.03cm 1.34cm 1.6cm 1.5cm,clip]{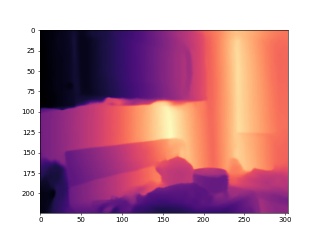} &
\includegraphics[width=\figSize,trim=2.03cm 1.34cm 1.6cm 1.5cm,clip]{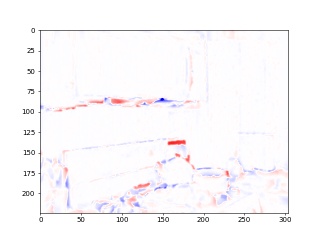} &
\includegraphics[width=\figSize,trim=2.03cm 1.34cm 1.6cm 1.5cm,clip]{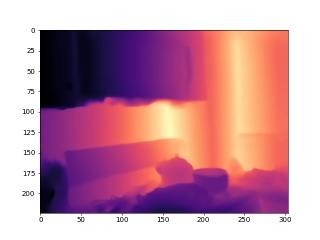} &
\includegraphics[width=\figSize,trim=2.03cm 1.34cm 1.6cm 1.5cm,clip]{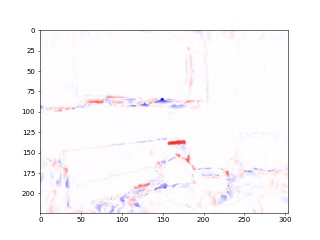} &
\includegraphics[width=\figSize,trim=2.03cm 1.34cm 1.6cm 1.5cm,clip]{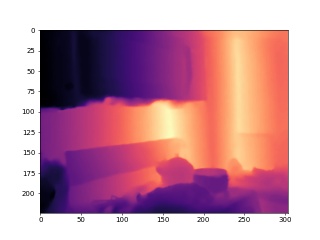} &
\includegraphics[width=\figSize,trim=2.03cm 1.34cm 1.6cm 1.5cm,clip]{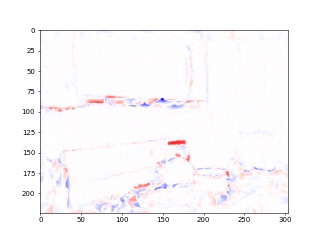} & 
\includegraphics[width=\figSize,trim=2.03cm 1.34cm 1.6cm 1.5cm,clip]{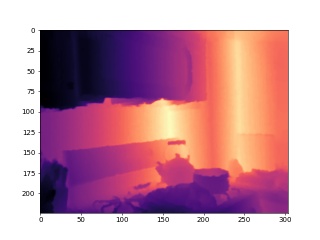} \\

\includegraphics[width=\figSize]{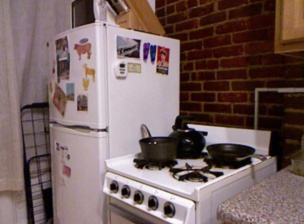} &
\includegraphics[width=\figSize,trim=2.03cm 1.34cm 1.6cm 1.5cm,clip]{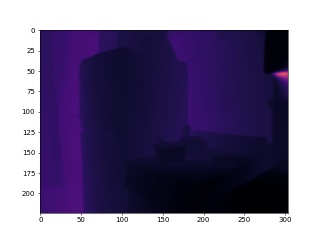} &
\includegraphics[width=\figSize,trim=2.03cm 1.34cm 1.6cm 1.5cm,clip]{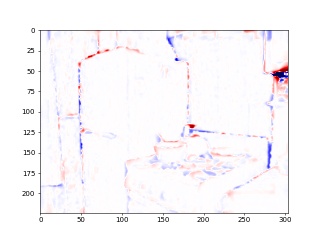} &
\includegraphics[width=\figSize,trim=2.03cm 1.34cm 1.6cm 1.5cm,clip]{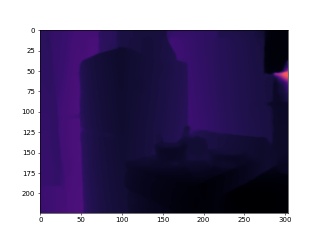} &
\includegraphics[width=\figSize,trim=2.03cm 1.34cm 1.6cm 1.5cm,clip]{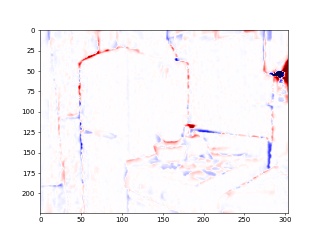} &
\includegraphics[width=\figSize,trim=2.03cm 1.34cm 1.6cm 1.5cm,clip]{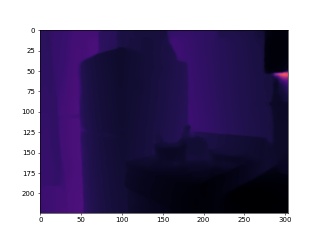} &
\includegraphics[width=\figSize,trim=2.03cm 1.34cm 1.6cm 1.5cm,clip]{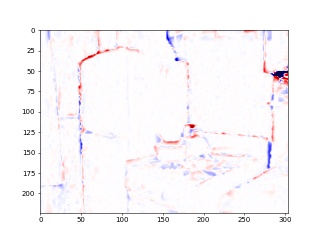} & 
\includegraphics[width=\figSize,trim=2.03cm 1.34cm 1.6cm 1.5cm,clip]{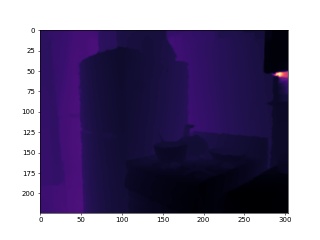} \\
\includegraphics[width=\figSize]{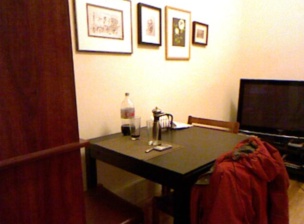} &
\includegraphics[width=\figSize,trim=2.03cm 1.34cm 1.6cm 1.5cm,clip]{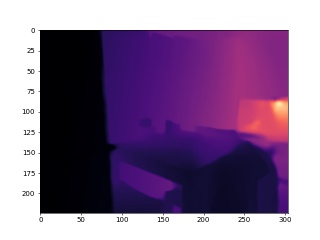} &
\includegraphics[width=\figSize,trim=2.03cm 1.34cm 1.6cm 1.5cm,clip]{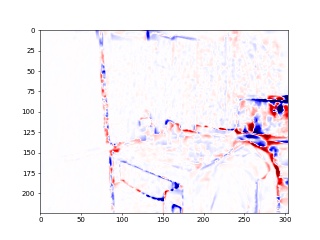} &
\includegraphics[width=\figSize,trim=2.03cm 1.34cm 1.6cm 1.5cm,clip]{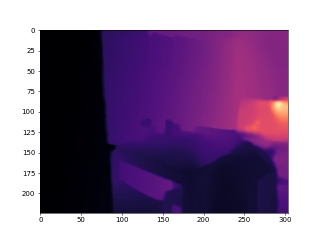} &
\includegraphics[width=\figSize,trim=2.03cm 1.34cm 1.6cm 1.5cm,clip]{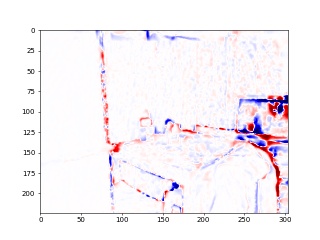} &
\includegraphics[width=\figSize,trim=2.03cm 1.34cm 1.6cm 1.5cm,clip]{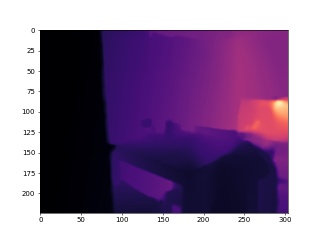} &
\includegraphics[width=\figSize,trim=2.03cm 1.34cm 1.6cm 1.5cm,clip]{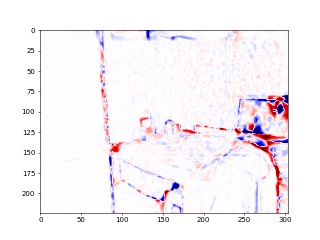} & 
\includegraphics[width=\figSize,trim=2.03cm 1.34cm 1.6cm 1.5cm,clip]{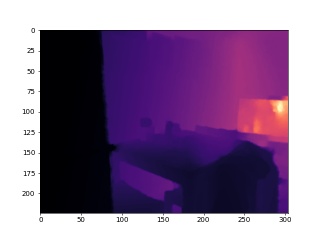} \\
%
\end{tabular}
\caption{Qualitative Results. We report, for several arbitrary frames in the test set of NYU-Depth v2, the predicted depth maps $\hat{D}$ and error maps $\hat{D}-D_{\rm GT}$ (range: $[-500, 500] mm$). Figure best viewed in color.}
\label{fig:results_q_NYU_sup}
\end{figure*}

We here extend the qualitative evaluation by reporting more images from the \sdsdata~(\cref{fig:results_q_SDS_sup}) and NYU-Depth v2  (\cref{fig:results_q_NYU_sup}) datasets, as processed by the reference methods in the main paper. These corroborate the evidence in the manuscript on the quality of our $\rm float32$ and mixed precision $\rm W_4 A_8$ models against NLSPN.

